\documentclass{article} 
\usepackage{iclr2026_conference,times}

\usepackage{amsmath,amsfonts,bm}

\def\eqref#1{equation~\ref{#1}}

\def\1{\bm{1}}

\DeclareMathAlphabet{\mathsfit}{\encodingdefault}{\sfdefault}{m}{sl}
\SetMathAlphabet{\mathsfit}{bold}{\encodingdefault}{\sfdefault}{bx}{n}

\DeclareMathOperator*{\argmax}{arg\,max}

\usepackage{hyperref}
\usepackage{url}

\title{ReDraft, Don't Just Distill: Reference-Driven Revision for Continual MLLM Post-Training}

\author{\textbf{Zhihao Zhang}$^{1,2}\thanks{{ }{ }Equal contributions.} \: \: $,
\ \textbf{Mingqi Wu}$^{1\,*}$, 
\ \textbf{Qiaole Dong}$^{3}$,
\ \textbf{Enyu Zhou}$^{1}$,
\ \textbf{Shuo Li}$^{1}$,
\ \textbf{Boyang Liu}$^{1}$,
\\
\textbf{Jiazheng Zhang}$^{1}$,
\ \textbf{Honglin Guo}$^{1}$, 
\ \textbf{Xin Guo}$^{1}$,
\ \textbf{Shaofan Liu}$^{1}$,
\ \textbf{Junzhe Wang}$^{1}$, 
\ \textbf{Dingwei Zhu}$^{1}$,
\\
\textbf{Minlong Peng}$^{3}$,
\ \textbf{Yuan Hua}$^{3}$,
\ \textbf{Zhiheng Xi}$^{1}$,
\ \textbf{Qi Zhang}$^{1,2}\thanks{{ }{ }Corresponding author.}\:  \:$,
\ \textbf{Tao Gui}$^{1\,\dagger}$,
\ \textbf{Xuanjing Huang}$^{1}$
\\
$^{1}$ Fudan University\quad$^{2}$ Shanghai Artificial Intelligence Laboratory\quad$^{3}$ Independent Researcher\\ 
\texttt{\{zhangzhihao19, qz, tgui\}@fudan.edu.cn}
}

\usepackage{iftex}
\usepackage{graphicx}     
\usepackage{wrapfig}      
\usepackage{booktabs}     
\usepackage{amsmath}      
\usepackage{amssymb}      
\usepackage{amsthm}       
\usepackage{makecell}     
\usepackage{xcolor}       
\usepackage[labelfont=bf,font=small]{caption}
\usepackage{microtype}
\usepackage{placeins}

\usepackage{hyperref}
\usepackage{url}
\usepackage{silence}
\usepackage{color}
\usepackage{subcaption}
\usepackage{enumitem}
\usepackage{multirow}
\usepackage{adjustbox} 
\usepackage{arydshln} 
\usepackage{array}
\usepackage{float}
\usepackage[most]{tcolorbox}

\newcommand{\method}{ReDraft}
\newcommand{\methodm}{\text{ReDraft}}

\newtheorem{assumption}{Assumption}

\newcommand{\Ccal}{\mathcal{C}}
\newcommand{\Dcal}{\mathcal{D}}

\iclrfinalcopy 
\begin{document}

\maketitle

\begin{abstract}

Continual post-training of large multimodal models should add new capabilities while preserving those from pre-training, and the two goals pull in opposite directions. SFT gives \emph{explicit target supervision} that learns a task from near-zero accuracy, but its off-policy targets move the model far enough to cause forgetting; on-policy methods such as RLVR and self-distillation preserve \emph{policy proximity} yet supply little signal when the policy cannot yet solve the task.
We introduce \method{} (Reference-Driven Revision and Fine-Tuning), which obtains both from the model's own failures: using an expert response only as a reference, it has the model revise its own incorrect rollout, keeps the revision only if a verifier accepts it, and fine-tunes on what survives. Each retained target is therefore explicit, yet still close to the current policy.
Across Counting, Clock Reading, and Jigsaw on Qwen2.5-VL-3B/7B, two of them with near-zero accuracy, \method{} gains $56.9$ points on the target task against SFT's $52.9$ while cutting prior-task loss from $16.6$ to $1.5$ points ($11.3\times$ less forgetting), and improves on OPSD along both axes ($19.3$ gain, $6.2$ loss). Data- and parameter-space analyses match the design: revised targets are more probable under the base model, and the updates they induce stay compact and follow SFT's direction more closely than OPSD's. Together, these results show that revising the model's own rollout rather than directly imitating an expert trajectory can reconcile cold-start acquisition with prior-capability retention.

\end{abstract}

\section{Introduction}
\label{sec:intro}

\begin{wrapfigure}{r}{0.48\textwidth}
  \vspace{-\intextsep}
  \centering
  \includegraphics[width=\linewidth]{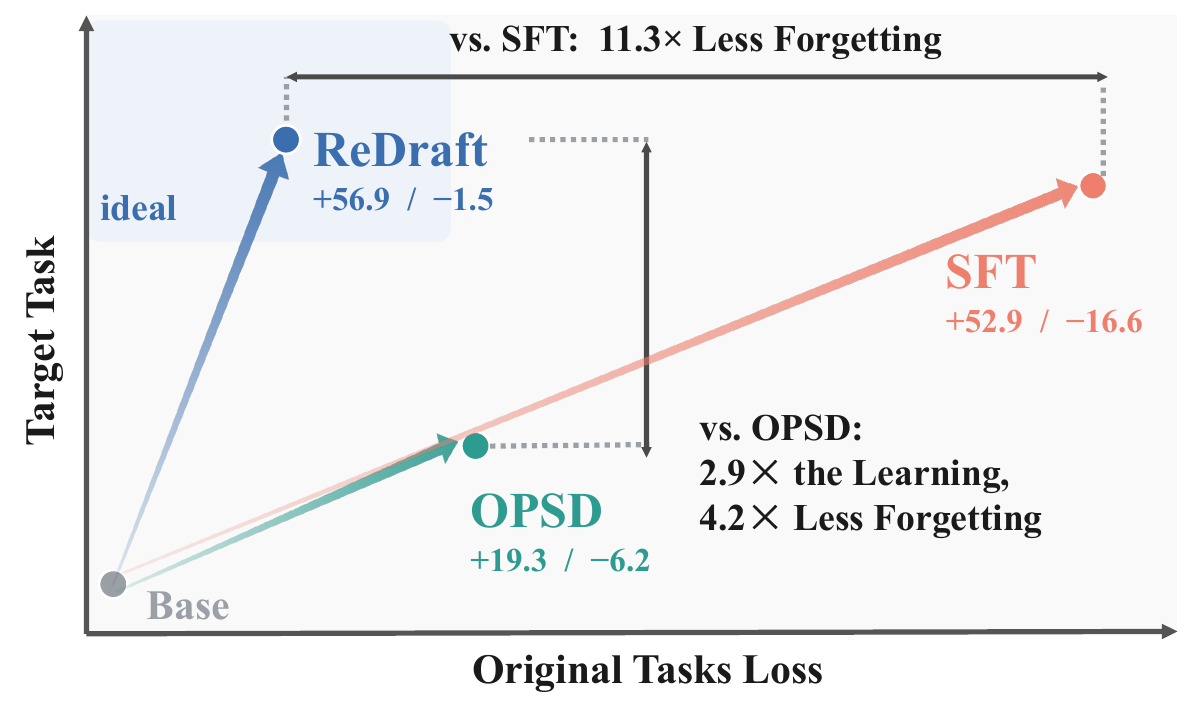}
  \vspace{-1.5\intextsep}
  \caption{Target-task gain against loss on the original tasks, averaged over all experiments. \method{} is in the upper-left region.}
  \vspace{-1.0\intextsep}
  \label{fig:teaser}
\end{wrapfigure}

Continual post-training aims to acquire new capabilities while preserving those learned during pre-training~\citep{kirkpatrick2017forgetting,luo2023forgetting}. These goals are often in tension. Effective acquisition, especially for tasks on which the base model has little or no prior competence, requires informative and explicit training targets. In contrast, retaining prior capabilities requires updates that remain close to the model's current behavior. We refer to these two properties as \emph{explicit target supervision} and \emph{policy proximity}, respectively. This raises a central question: \emph{can we provide explicit supervision for cold-start learning while maintaining proximity to the current policy?}

Existing post-training approaches largely emphasize one of these properties at the expense of the other. Supervised Fine-Tuning (SFT) provides strong \emph{explicit target supervision}: given an expert response $y^\star$, it specifies the desired output at every token, enabling rapid learning even when base-model accuracy is near zero~\citep{wei2022flan,chu2025sftrl}. Because expert responses are generated independently of the current policy, however, fitting these off-policy targets may induce large behavioral changes and catastrophic forgetting~\citep{luo2023forgetting,zhang2026rft}. Reinforcement Learning with Verifiable Rewards (RLVR) takes the opposite approach, sampling responses from the current policy and scoring them with a verifier; this on-policy training improves \emph{policy proximity} and helps preserve prior capabilities~\citep{lambert2024tulu3,shenfeld2025razor,chen2026retaining}. It therefore refines capabilities that the policy can already partially express~\citep{wu2026reasoning}, but supplies little supervision on a task where almost no sampled response is correct.


Recently, on-policy self-distillation methods such as SDFT~\citep{shenfeld2026selfdistillation} and OPSD~\citep{zhao2026opsd} seek to bridge this gap by combining an expert response with trajectories sampled from the current policy~\citep{hubotter2026rlselfdistillation}. Their correction remains largely implicit, conveyed through a per-token divergence between the reference-conditioned teacher and the student~\citep{lu2025onpolicydistillation,hou2026uniopd,lv2024wasserstein}: because the two differ only in the reference prefix, the largest divergences often fall on stylistic or pivot tokens rather than on the decisions that settle the task, yielding a dense but potentially misaligned training signal. Per-token clipping can help constrain the resulting policy updates~\citep{zhao2026opsd,gu2024minillm,nrehiew2026distributional}. Even when a divergence reflects a genuine error, matching the teacher at that position may not suffice to redirect a continuation conditioned on the student's erroneous prefix, so the error may recur later in the response~\citep{jiang2026trajectories,fu2026revisiting}. This limitation becomes more pronounced at cold start, when a rollout may contain multiple interdependent errors: self-distillation maintains \emph{policy proximity}, but does not directly provide an explicit corrected trajectory to learn from.

\begin{figure}[t]
  \centering
  \vspace{-0.5\intextsep}
  \includegraphics[width=\textwidth]{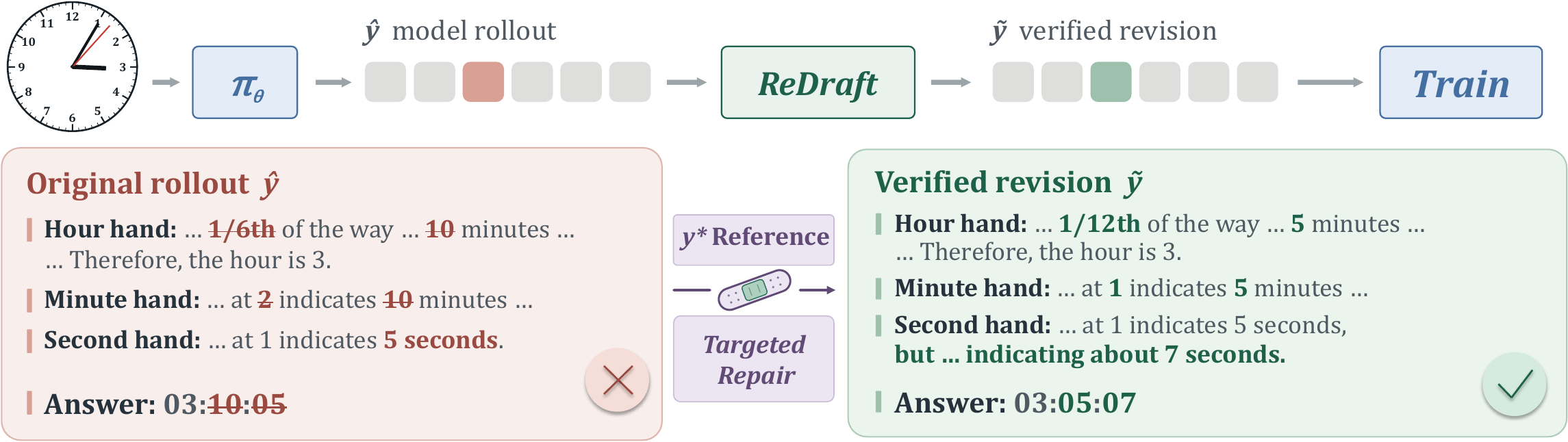}
  \vspace{-1\intextsep}
  \caption{\textbf{Overview of \method{}}. Each round samples a response, revises incorrect responses using an expert response as a reference, verifies the revisions, and fine-tunes on the resulting targets. By explicitly incorporating the necessary corrections, \method{} combines \emph{explicit target supervision} with \emph{policy proximity}.}
  \vspace{-1\intextsep}
  \label{fig:overview}
\end{figure}

We propose \method{} (Reference-Driven Revision and Fine-Tuning) to address this limitation by revising the student's rollout into an explicit training target. For each prompt, \method{} first samples a response from the current policy. Correct responses are kept unchanged, while incorrect responses are revised by the model itself, using an expert response $y^\star$ as a reference. Each revision is then verified, and only successful revisions are used for fine-tuning (Figure~\ref{fig:overview}). Rather than providing token-level corrections along the original, potentially incorrect trajectory, \method{} externalizes the correction as a complete revised response. Because the revision starts from an on-policy rollout and is generated by the policy itself, the resulting target remains close to the current policy; we therefore call it \emph{self-revised supervision}. This construction combines \emph{explicit target supervision} with \emph{policy proximity}: an incorrect rollout can be converted into a positive training target without requiring the current policy to first discover a correct solution path.

We evaluate \method{} on Counting~\citep{deitke2025molmo}, Clock Reading~\citep{fu2025clock}, and Jigsaw~\citep{wang2025jigsawr1,lin2014coco} using Qwen2.5-VL-3B and 7B~\citep{bai2025qwen25vl}, where Clock Reading and Jigsaw are genuine cold-start settings with near-zero base-model accuracy.  With retention measured over 8 benchmarks of prior capabilities~(Section~\ref{sec:results}), \method{} exceeds SFT's target-task gain with $11.3\times$ less forgetting, and improves on OPSD along both axes, with nearly three times its gain at under a quarter of its forgetting (Figure~\ref{fig:teaser}); the advantage persists under mixed and sequential multi-task training. Further analyses link the two halves of this result to the two design properties~(Section~\ref{sec:ablation-weightspace}): verified self-revisions have lower perplexity under the base model than external SFT targets, consistent with greater \emph{policy proximity}, while in parameter space \method{} produces compact updates that follow SFT's direction more closely than OPSD's, as \emph{explicit target supervision} predicts. The complementary failure is visible in OPSD, whose token-level credit often falls on incidental reasoning or stylistic tokens rather than the decisions that determine task success.
Finally, when we withhold the target answer from the revision instruction, repairs fall from above $94\%$ at 32B and 72B to under $3\%$ at 3B, so the ability to revise has a scale threshold of its own~(Section~\ref{sec:revision-capability}). We therefore propose \method{}$^{*}$: where a model cannot repair its own response, a stronger expert can supply the revision instead of the trajectory, editing what requires correction and leaving the rest of the rollout intact; training on these targets retains most of \method{}'s advantage over SFT.



Our contributions are four-fold:
\begin{itemize}[leftmargin=1.5em,itemsep=2pt,topsep=2pt]
\item We propose \method{}, a post-training method built on \emph{self-revised supervision}, which converts failed on-policy rollouts into self-revised, verifier-approved targets, combining \emph{explicit target supervision} with \emph{policy proximity}.
\item Across three tasks and two model scales, we show that \method{} matches or exceeds SFT-level acquisition while achieving $11.3\times$ less forgetting, and consistently outperforms OPSD in both target-task acquisition and prior-task retention.
\item Data- and parameter-space analyses attribute this advantage to the two design properties: revised targets remain close to the current policy, while the updates they induce follow SFT's direction more closely than OPSD's.
\item We show that the model's   ability to revise its own rollout scales with model size. By leveraging the revision capability of an external expert model rather than just distilling its trajectory, a 3B model can also acquire a new task while retaining more prior capability than SFT.
\end{itemize}

\section{Related Work}
\label{sec:related-work}

\paragraph{Off-policy and On-policy Post-training.} Supervised fine-tuning (SFT) adapts language and multimodal models from curated demonstrations, providing explicit token-level imitation targets~\citep{wei2022flan,yu2024metamath,paster2024openwebmath}. Because these trajectories are fixed independently of the learner, SFT is off-policy and can suffer from exposure bias, memorization, and distribution shift~\citep{ross2011dagger,chu2025sftrl}. A parallel line of work keeps these targets but reshapes the objective, clipping or reweighting per-token updates against a reference policy to limit policy drift~\citep{zhu2026psft,zhu2026asft,liu2026profit}; \method{} is complementary, leaving the objective intact and replacing the targets themselves with verified self-revised responses. Reinforcement learning instead optimizes sampled trajectories from outcome or preference feedback, yielding strong reasoning and out-of-distribution generalization~\citep{schulman2017ppo,shao2024deepseekmath,guo2025deepseekr1,huan2025transferability}, and its on-policy data interfere less with prior capabilities~\citep{shenfeld2025razor,lai2025rft,zhang2026rft,chen2026retaining}. Both benefits presuppose informative rewards: when virtually every rollout fails, verifier feedback supplies no positive trajectory. We target this cold-start regime, where SFT acquires the skill but forgets, and RLVR forgets little but cannot bootstrap it.

\paragraph{On-policy Self-distillation.} Knowledge distillation conventionally transfers a separate teacher's output distribution to a student~\citep{hinton2015distilling,kim2016sequence}, while on-policy distillation evaluates teacher guidance along trajectories sampled by the student, reducing train--test mismatch~\citep{agarwal2024gkd,xu2024speculative,lu2025onpolicydistillation}. Context-distillation methods further use privileged prompts or demonstrations to create a stronger conditional version of the same model~\citep{bai2022constitutional,snell2022learning}. SDFT applies this idea to continual learning by distilling a demonstration-conditioned teacher on student rollouts, and OPSD conditions the teacher on ground-truth reasoning solutions to improve reasoning~\citep{shenfeld2026selfdistillation,zhao2026opsd,yuan2026visionopd}. Both obtain dense token-level supervision from on-policy data without a separate teacher. Their correction nevertheless remains implicit in teacher--student divergence and need not concentrate on task-critical tokens. 
\method{} instead makes this correction explicit by having the policy generate a verified revised response, retaining the data-distribution benefits of on-policy distillation while avoiding reliance on implicit KL-based credit assignment.

\paragraph{Catastrophic Forgetting.} Catastrophic forgetting was first identified in sequential neural-network training~\citep{mccloskey1989catastrophic,ratcliff1990connectionist}. Classical remedies regularize parameters important to old tasks~\citep{kirkpatrick2017forgetting,zenke2017synaptic,li2018learning}, replay prior examples~\citep{shin2017generative,rebuffi2017icarl,chaudhry2019efficient}, or allocate task-specific parameters~\citep{rusu2016progressive,serra2018hard,zhang2024linguistic}. These approaches are difficult to apply to foundation-model post-training because pre-training data are commonly unavailable, extra modules are costly, and rigid constraints can impede acquisition. Recent work instead finds that online RL and reinforcement fine-tuning forget less than SFT
~\citep{shenfeld2025razor,zhang2026rft,chen2026retaining}. \method{} follows this data-centric direction but targets cold start: verified self-revised targets reduce forgetting without replay, frozen parameters, or waiting for rare successful rollouts.

\paragraph{Self-training and Expert-revision.} LLM self-training methods generate their own instructions, rationales, or responses and reuse selected outputs for learning~\citep{wang2023selfinstruct,chen2024selfplay,yuan2024selfrewarding}. STaR~\citep{zelikman2022star} and ReST~\citep{gulcehre2023reinforced}, for example, condition rationale generation on answers or hints, filter trajectories by correctness or reward, and fine-tune on the retained sequences; related self-refinement and reflection methods iteratively revise responses using model or environment feedback~\citep{madaan2023selfrefine,shinn2023reflexion,gou2024critic}.
Three concurrent methods build targets from the model's own generations as we do, each differing in one respect: ReGFT supplies a partial reference and lets the model complete its own trace, so no failed rollout is edited~\citep{wu2026regft}; SPoT has an external oracle do the editing and enforces proximity by surface-form overlap~\citep{lin2026surgical}; and SD-Zero conditions revision on the binary reward alone, with no reference~\citep{he2026sdzero}. \method{} instead edits the policy's own failed response against an expert reference, keeps it only if verified, and regenerates this corpus as training proceeds. Cold-start supervision is therefore explicit and drawn from the model's own trajectories.

\section{Background and Method}
\label{sec:background}

\subsection{Setup}
\label{sec:setup}

\paragraph{Problem setting.} We post-train a pretrained policy $\pi_{0}$ on a
target task $\mathcal{T}$ on which it has little or no initial competence, so
the accuracy of its sampled responses is near zero and supervision cannot be obtained by
filtering them. We therefore assume two sources of external information: a set
of demonstrations $\Dcal=\{(x_i,y_i^{\star})\}_{i=1}^{N}$ that pairs every
prompt with an expert reference trajectory $y^{\star}$, and a verifier
$v(y,y^{\star})\in\{0,1\}$ that judges whether a complete response is correct.
Training starts from $\pi_{0}$ and updates the parameters $\theta$; we denote
the policy at the current training step by $\pi_\theta$.

\paragraph{Evaluation axes.} Continual post-training is judged on two axes at
once: the \emph{gain} on the target task and the \emph{forgetting} on
$\mathcal{P}$, a benchmark suite assessing the base model's prior capabilities,
\begin{equation}
  G(\theta)=\mathrm{Acc}_{\mathcal{T}}(\pi_\theta)-\mathrm{Acc}_{\mathcal{T}}(\pi_{0}),
  \qquad
  F(\theta)=\mathrm{Acc}_{\mathcal{P}}(\pi_{0})-\mathrm{Acc}_{\mathcal{P}}(\pi_\theta),
  \label{eq:gain-loss}
\end{equation}
where $\mathrm{Acc}$ is accuracy on the corresponding evaluation set.

\subsection{Learning from Demonstrations}
\label{sec:twoways}

\paragraph{Supervised fine-tuning (SFT).} Supervised fine-tuning imitates the expert trajectory
directly:
\begin{equation}
  \mathcal{L}_{\mathrm{SFT}}(\theta)
  = \mathbb{E}_{\Dcal}\big[-\log\pi_\theta(y^{\star}\mid x)\big].
  \label{eq:sft}
\end{equation}

\paragraph{On-policy self-distillation.} Rather than fitting $y^\star$ itself, SDFT~\citep{shenfeld2026selfdistillation} and OPSD~\citep{zhao2026opsd} place it in the teacher context and distill the reference-conditioned model into the unconditioned one along the student's own rollout. On a rollout $\hat y$, let $q_{\theta,n}=\pi_\theta(\cdot\mid x,y^\star,\hat y_{<n})$ and $p_{\theta,n}=\pi_\theta(\cdot\mid x,\hat y_{<n})$; with forward KL, the objective is
\begin{equation}
  \mathcal{L}_{\mathrm{OPD}}(\theta)
  = \mathbb{E}_{\Dcal}\,
    \mathbb{E}_{\hat{y}\sim\pi_\theta(\cdot\mid x)}
    \frac{1}{|\hat{y}|}\sum_{n=1}^{|\hat{y}|}
    \mathrm{KL}(q_{\theta,n}\|p_{\theta,n}),
  \label{eq:opd}
\end{equation}
with gradients taken through the student branch only.

\subsection{Self-Revised Supervision}
\label{sec:selfrevised}

\paragraph{An ideal training target.} Our goal is to improve target-task accuracy
while limiting changes to the current policy. For a fixed training pair
$(x,y^\star)$, we express this trade-off as maximizing verifier reward with a
KL penalty~\citep{schulman2015trpo}:
\begin{equation}
  \pi^{\star}_{\beta}
  =\argmax_{\pi}\;
    \mathbb{E}_{y\sim\pi(\cdot\mid x)}\big[v(y,y^{\star})\big]
    -\beta\,D_{\mathrm{KL}}\big(\pi(\cdot\mid x)\,\big\|\,\pi_\theta(\cdot\mid x)\big).
  \label{eq:trust-region}
\end{equation}
The solution for $\beta>0$ is the tilted distribution
$\pi^{\star}_{\beta}(y\mid x)\propto\pi_\theta(y\mid x)\exp\!\big(v(y,y^{\star})/\beta\big)$~\citep{korbak2022klbayes,rafailov2023dpo}.
For a binary verifier, taking $\beta\to0^{+}$ gives the ideal target
(Appendix~\ref{app:limit}):
\begin{equation}
  \pi^{\star}(y\mid x)
  =\frac{\pi_\theta(y\mid x)\,\mathbf{1}\{y\in\Ccal(x)\}}{Z_\theta(x)},
  \qquad
  Z_\theta(x)=\Pr_{y\sim\pi_\theta(\cdot\mid x)}\big[y\in\Ccal(x)\big]>0,
  \label{eq:ideal-target}
\end{equation}
where $\Ccal(x)=\{y:v(y,y^{\star})=1\}$ is the set of responses accepted by
the verifier. The indicator excludes incorrect
responses, so a sample from this target is a complete, verified response for
\emph{explicit target supervision}. Among correct responses, the $\pi_\theta$
factor preserves the model's relative preferences rather than concentrating
all mass on an external reference. The resulting target is the closest
distribution to $\pi_\theta$ under the KL penalty above subject to correctness,
providing a precise notion of \emph{policy proximity}.

\paragraph{Limitations of existing objectives.} RLVR and rejection sampling draw from
this target but require $1/Z_\theta(x)$ samples per accepted response on
average, which becomes costly when $Z_\theta(x)\approx0$. SFT instead fits $\delta_{y^\star}$,
the distribution assigning all probability to the expert response. This
provides a correct target but discards the effects of $\pi_\theta$ weighting in
Equation~\ref{eq:ideal-target}.

Self-distillation uses the reference-conditioned teacher as a proxy for this
ideal target~\citep{shenfeld2026selfdistillation}. To compare their
distributions, we write the teacher as a reweighted current policy:
\begin{equation}
  \pi_\theta(y\mid x,y^\star)=\pi_\theta(y\mid x)\,w_\theta(y),
  \qquad
  w_\theta(y)=\frac{\pi_\theta(y\mid x,y^\star)}{\pi_\theta(y\mid x)}.
  \label{eq:opd-target}
\end{equation}
This identity assumes shared support and requires no further normalization
(Appendix~\ref{app:cost}). Unlike the correctness indicator in
Equation~\ref{eq:ideal-target}, the unverified weight $w_\theta(y)$ need not
vanish on incorrect responses.
Thus Equation~\ref{eq:opd} matches teacher probabilities along the student's
rollout without constructing an explicit, verified correction.

Consequently, a large token-level KL need not identify a task-critical
error: Equation~\ref{eq:opd} penalizes any disagreement between the teacher
and student, including differences in style. OPSD reports stylistic tokens
contributing $6$--$15\times$ more KL than mathematical ones and uses pointwise
clipping to limit their influence~\citep{zhao2026opsd}. Clipping controls
large contributions, but does not identify the tokens that need correction
or provide a corrected continuation.


\begin{wrapfigure}{r}{0.44\textwidth}
  \vspace{-\intextsep}
  \centering
  \includegraphics[width=\linewidth]{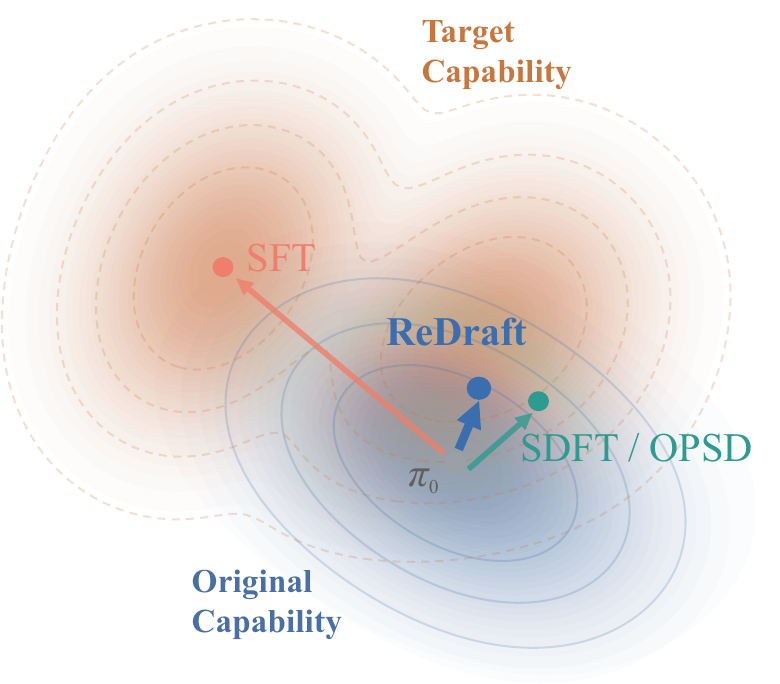}
  \caption{\textbf{Schematic of where each objective moves the policy.} From
  $\pi_0$, SFT reaches the target capability but leaves the original one;
  SDFT/OPSD keeps the original but barely approaches the target; \method{}
  reaches the target without leaving the original.}
  \vspace{-1.4\intextsep}
  \label{fig:landscape}
\end{wrapfigure}

\paragraph{Revision proposal.} \method{} replaces implicit token-level correction
with explicit revision and verification of the model's own rollout. It first
samples $\hat y\sim\pi_\theta(\cdot\mid x)$ and sets $\tilde y=\hat y$ if the
verifier accepts it. Otherwise, the same model generates $\tilde y$ from
$(x,y^\star,\hat y)$, with instructions to preserve valid content and correct
errors. We denote this keep-or-revise distribution by
$R_\theta(\cdot\mid x,y^\star,\hat y)$.

\begin{assumption}[Correction competence]
\label{as:correction}
For a prompt with $Z_\theta(x)<1$, let
$\tilde y\sim R_\theta(\cdot\mid x,y^\star,\hat y)$. The probability of
correcting a failed rollout is positive:
\begin{equation*}
  r_\theta(x)
  =\Pr\!\left[\tilde y\in\Ccal(x)\,\middle|\,\hat y\notin\Ccal(x)\right]>0,
  \ \hat y\sim\pi_\theta(\cdot\mid x).
\end{equation*}
\end{assumption}

\noindent For each fixed $(x,y^\star)$, averaging over initial rollouts gives
the candidate distribution below, with the reference dependence suppressed:
\begin{equation}
  \rho_\theta(y\mid x)
  =\sum_{\hat y}\pi_\theta(\hat y\mid x)\,R_\theta(y\mid x,y^\star,\hat y).
  \label{eq:revision-proposal}
\end{equation}
Among candidates that pass verification, let $\pi_{\methodm}(y\mid x)$
denote the probability that the retained response equals $y$. Conditional
probability gives
\begin{equation}
  \pi_{\methodm}(y\mid x)
  =\frac{\rho_\theta(y\mid x)\,\mathbf{1}\{y\in\Ccal(x)\}}{S_\theta(x)},
  \qquad
  S_\theta(x)=Z_\theta(x)+\big(1-Z_\theta(x)\big)r_\theta(x)\;\ge\;Z_\theta(x).
  \label{eq:redraft-target}
\end{equation}
Here $S_\theta(x)=\Pr_{\tilde y\sim\rho_\theta}[\tilde y\in\Ccal(x)]>0$ is the
candidate acceptance rate, and it describes
the training targets produced by our procedure. 
At a fixed number of initial rollouts,
revision yields at least as many accepted targets as rejection sampling in
expectation: at cold start $Z_\theta(x)\!\approx\!0$, so
$S_\theta(x)\!\approx\!r_\theta(x)$.
Appendix~\ref{app:yield} reports both rates for every task and scale:
on the cold-start tasks, initial acceptance is low while repair rates exceed
$56\%$. Examples of revised responses appear in Appendix~\ref{app:demos}.

\paragraph{ReDraft Supervised Fine-tuning.} Each retained response becomes a training
pair $(x,\tilde y)$, with $\tilde y$ drawn from $\pi_{\methodm}(\cdot\mid x)$.
The learner predicts this target from $x$ alone, without the reference or
initial rollout. Holding the collection process fixed, the expected loss is
\begin{equation}
  \mathcal{L}_{\methodm}(\theta)
  =\mathbb{E}_{\Dcal}\Big[S_\theta(x)\,
    \mathbb{E}_{\tilde y\sim\pi_{\methodm}(\cdot\mid x)}
    \big[-\log\pi_\theta(\tilde y\mid x)\big]\Big].
  \label{eq:redraft}
\end{equation}
The factor $S_\theta(x)$ reflects how often a prompt supplies a retained
target, not an additional weight applied to accepted examples.

\noindent Every retained target in Equation~\ref{eq:redraft} is verifier-approved.
Edited spans provide explicit supervision, while unchanged spans preserve
the model's own trajectory and encourage policy proximity.

\begin{wraptable}{r}{0.47\textwidth}
  \vspace{-\intextsep}
  \centering
  {\footnotesize
   \setlength{\tabcolsep}{3pt}
   \renewcommand{\arraystretch}{1.12}
   \begin{tabular}{@{}lcccc@{}}
     \toprule
     & RLVR & SFT & OPSD & \method{} \\
     \midrule
     Policy proximity        & \checkmark & ---        & \checkmark & \checkmark \\
     Dense signal            & ---        & \checkmark & \checkmark & \checkmark \\
     Explicit target & \checkmark & \checkmark & ---        & \checkmark \\
     Usable at zero start    & ---        & \checkmark & ---        & \checkmark \\
     \bottomrule
   \end{tabular}}
  \captionof{table}{\textbf{What each supervision method provides.}
  \method{} pairs a policy-proximal target with dense and explicitly correct
  target supervision.}
  \label{tab:paradigms}
  \vspace{-0.5\intextsep}
\end{wraptable}

\paragraph{Summary.} Table~\ref{tab:paradigms} compares the objectives on the four properties that decide the cold-start case: combines SFT's explicit targets supervision with RLVR and OPSD's policy proximity, and is the only column that supplies all four. Figure~\ref{fig:landscape} gives the same picture geometrically: the model moves toward the target capability without imitating SFT's distant expert trajectory or relying on OPSD's implicit KL allocation. This predicts SFT-like acquisition with the retention benefit of on-policy data.

\section{Experimental Setup}
\label{sec:experiments}
\textbf{Task Acquisition.} We validate our hypothesis on three canonical vision-language tasks. \textit{Jigsaw} and \textit{Clock Reading} are our cold-start settings, as base model's accuracy is near zero:
\begin{itemize}[leftmargin=1em,itemsep=2pt,topsep=2pt]
    \item \textbf{Counting.} Given an image, the model is asked to count the instances of a specified object category, such as cats, birds, or people. The base model is already partly competent here, so this task measures the refinement of an existing skill.
    \item \textbf{Jigsaw.} A real-world image is partitioned into four patches by a $2\times2$ grid and randomly shuffled; the model must recover the correct ordering of the patches.
    \item \textbf{Analog Clock Reading.} Given a synthetically rendered image of an analog clock, the model is asked to report the time it displays.
\end{itemize}

\textbf{Dataset Construction.} Each target task draws on an existing image
corpus: for Counting we sample $2$k images from
PixMo-Count~\citep{deitke2025molmo}; for Clock Reading we take $10$k standard clock images from Analog Clocks Combinations~\citep{fu2025clock}; and for Jigsaw we sample $10$k
$640\times480$ images from COCO~\citep{lin2014coco}, cut each
into four equal patches, and shuffle them at random. Every prompt is then
paired with an expert response $y^\star$ generated by
GPT-5.5~\citep{openai2026gpt55}, and the same responses serve every objective
that needs one: they are the SFT targets, the reference \method{} revises
against, and the response on which OPSD conditions its teacher.

\textbf{MLLMs.} We employ Qwen2.5-VL-3B~\citep{bai2025qwen25vl} and 7B as our MLLMs due to their strong performance on vision-language understanding and support of native-resolution input.  

\textbf{Evaluation.}
We evaluate the post-trained model not only on novel tasks, but also on $4$  representative capability axes of prior knowledge:
\begin{itemize}[leftmargin=1em,itemsep=2pt,topsep=2pt]
    \item \textbf{OCR, chart \& document understanding.} AI2D \citep{kembhavi2016diagram}, DocVQA \citep{mathew2021docvqa}, InfoVQA \citep{mathew2022infovqa}, and ChartQA \citep{masry2022chartqa} probe the ability of MLLMs to read and reason over scanned documents, forms, and scientific plots.
    \item \textbf{General VQA.} MME \citep{fu2025mme} and MMStar \citep{chen2024mmstar} cover visual reasoning, spatial relations, and multimodal commonsense.
    \item \textbf{Real-world spatial understanding.} RealWorldQA \citep{xai2024realworldqa} probes spatial relations and physical commonsense in everyday photographs.
    \item \textbf{Science.} ScienceQA \citep{lu2022scienceqa} focuses on diagrams and textbook illustrations that require domain knowledge.
\end{itemize}

\textbf{Hyper-parameter setup.}
We implement all experiments on top of the TRL framework~\citep{vonwerra2020trl}.
Across all runs, we use a learning rate of $1\times10^{-5}$ and a batch size of $32$.
We set \texttt{max\_length} to either $4,096$ or $2,048$, depending on the length of the training corpus.
For the OPSD experiments, we set \texttt{token\_clip} to $0.0$, which disables element-wise clipping and therefore optimizes the exact forward KL, and set the temperature to $1.1$.
At evaluation time on the target tasks, we decode with a temperature of $0.7$ and sample $8$ responses per question, reporting the $avg$@8.

\section{Results and Analysis}
\label{sec:results}

\subsection{Main Results}
\label{sec:main-results}

\begin{figure}[t]
  \vspace{-\intextsep}
  \centering
  \includegraphics[width=\linewidth]{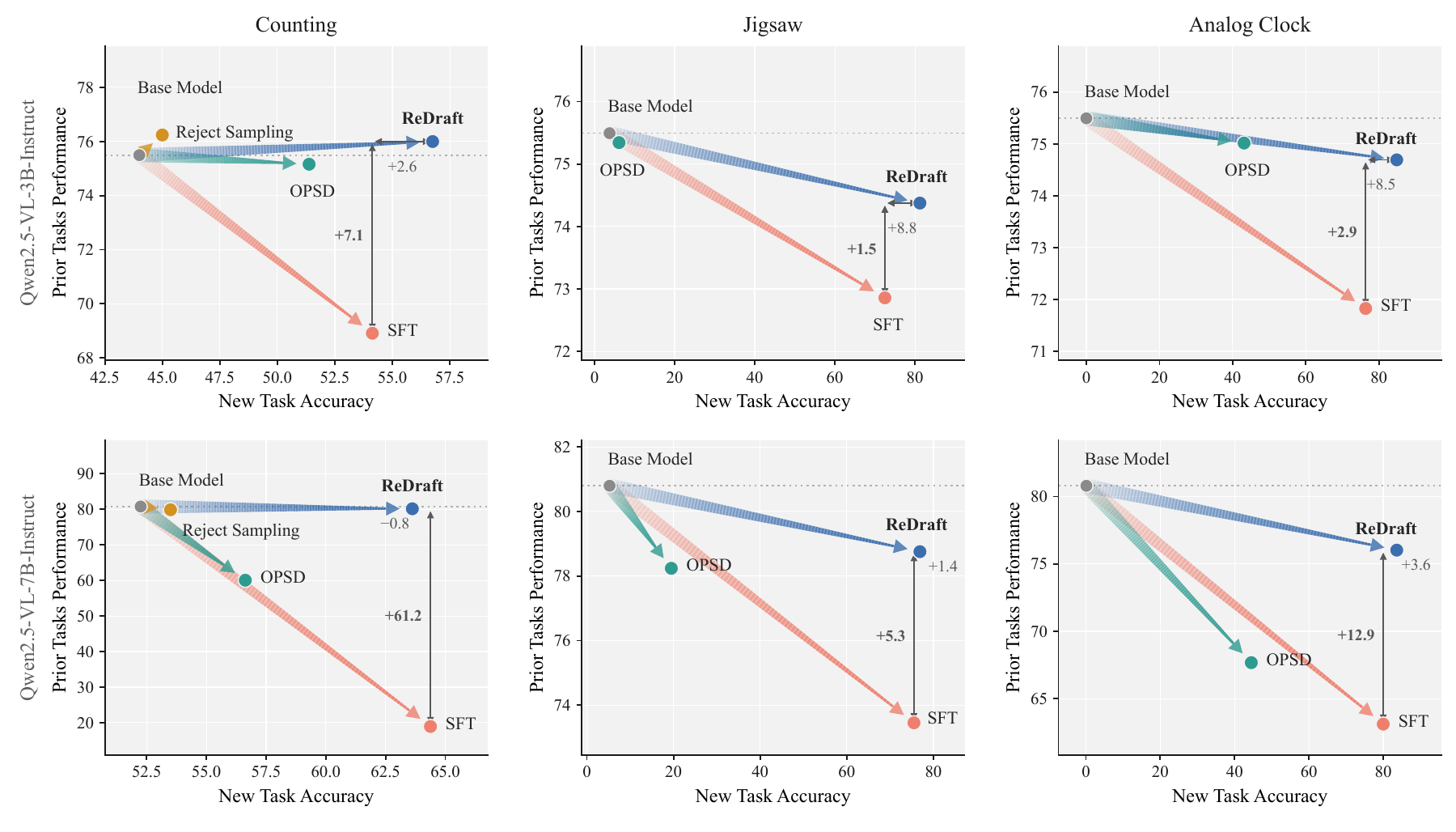}
  \caption{\textbf{Target-task acquisition vs. retention of prior
  capabilities.} Rows show Qwen2.5-VL-3B/7B; arrows originate at the base
  model. Reject Sampling is evaluated only on Counting,
  where the base policy already has substantial competence. Brackets show the
  \method{}--SFT difference; higher is better on both axes.}
  \label{fig:main-results}
\end{figure}

\paragraph{Evaluation protocol.} We evaluate Qwen2.5-VL-3B and 7B on
Counting, Jigsaw, and Analog Clock Reading, with the latter two constituting
cold-start tasks: base-model accuracy is at most $5.25\%$. Retention is the
macro-average over $8$ prior benchmarks after normalizing MME by its
maximum score $2,800$. 
Gain and loss are measured from the corresponding base model and averaged over all
experiments.

\paragraph{\method{} preserves SFT-level acquisition while sharply reducing
forgetting.} \method{} attains a higher mean target-task gain than SFT
($56.9$ versus $52.9$ points) while reducing prior-task loss from $16.6$ to
$1.5$ points, or $11.3\times$ less forgetting. It leads SFT on target gain in
most settings and forgets less in every one (Figure~\ref{fig:main-results}).
Table~\ref{tab:detailed-results} in Appendix~\ref{app:detailed-results}
reports the detailed results of every experiment. \method{} therefore
matches or slightly improves target-task learning while retaining
substantially more original capability.

\paragraph{The advantage is largest when genuine cold start is required.}
Across Jigsaw and Analog Clock Reading, \method{} gains $79.4$ target-accuracy
points on average, vs. $73.8$ for SFT and $26.0$ for OPSD, while losing
only $2.2$ prior-task points, versus $7.8$ and $4.1$. The contrast with OPSD
is sharpest at 3B: on Jigsaw it barely leaves its base accuracy, ending at
$6.1\%$ target accuracy for a gain of $2.4$ points against \method{}'s
$77.5$, and on Clock Reading it recovers about half of \method{}'s gain
($43.1$ versus $84.9$). Across all settings, \method{} dominates OPSD and forgets less than SFT in
every pair.

\paragraph{RLVR and rejection sampling depend on successful rollouts.}
We further evaluate the two verifier-based alternatives to expert supervision, GRPO and Reject Sampling (RS).
GRPO improves accuracy only on 7B Jigsaw, where correct orderings are
occasionally sampled. In the other cold-start settings, correct responses
remain scarce and reward increases mainly reflect the format term rather
than task accuracy (Appendix~\ref{app:rlvr-coldstart}).
RS also relies on already-correct samples.
We report RS on Counting, where base accuracy is $44.0\%/52.3\%$ for 3B/7B
and positive rollouts are readily available. Even there, RS gains only
$1.0/1.3$ points, compared with \method{}'s $12.8/11.4$.
RS learns only from already-correct rollouts, whereas revision also converts
failed attempts into positive training targets.

\subsection{Is OPSD simply under-trained?}

\begin{wraptable}{r}{0.45\textwidth}
  \vspace{-\intextsep}
  \centering
  \small
  \setlength{\tabcolsep}{4pt}
  \caption{\textbf{Extending OPSD does not close the gap.} Target-task gain in
  percentage points, averaged over 3B and 7B.}
  \label{tab:opsd-extension}
  \begin{tabular}{@{}lccc@{}}
    \toprule
    Task      & \method{} & \multicolumn{2}{c}{OPSD} \\
    \cmidrule(lr){3-4}
              &           & standard & extended \\
    \midrule
    Counting      & $\mathbf{+12.1}$ & $+5.9$  & $+8.1$  \\
    Jigsaw        & $\mathbf{+74.6}$ & $+8.3$  & $+9.6$  \\
    Clock Reading & $\mathbf{+84.2}$ & $+43.8$ & $+52.5$ \\
    \midrule
    Mean gain     & $\mathbf{+56.9}$ & $+19.3$ & $+23.4$ \\
    Mean steps    & 367              & 449     & 762     \\
    \bottomrule
  \end{tabular}
  \vspace{-0.5\intextsep}
\end{wraptable}

In Figure~\ref{fig:main-results}, OPSD moves in a trade-off direction similar to \method{}'s in some settings, notably 3B Counting and Clock Reading, raising the possibility that its weaker acquisition is merely an optimization-budget effect. Therefore, we extend OPSD to more steps and, to favor the baseline, report its best-target checkpoint (Table~\ref{tab:opsd-extension}). The extra budget changes little: $70\%$ more steps buy $4.1$ points of gain ($19.3\!\rightarrow\!23.4$), and even on Clock Reading, where the longer budget helps most ($+8.7$), OPSD remains $31.7$ points behind \method{}, which reaches $56.9$ in fewer steps. Longer training therefore does not break the observed plateau. This is consistent with an objective-limited failure: on a new task, OPSD provides no sufficiently explicit corrective target for the missing behavior, so additional steps carry little usable signal. Table~\ref{tab:opsd-detail} in Appendix~\ref{app:opsd-extension} reports both budgets for each task and model scale.

\subsection{Multi-Task Training}
\label{sec:mixed-training}

\begin{figure}[ht!]
  \centering
  \includegraphics[width=\linewidth]{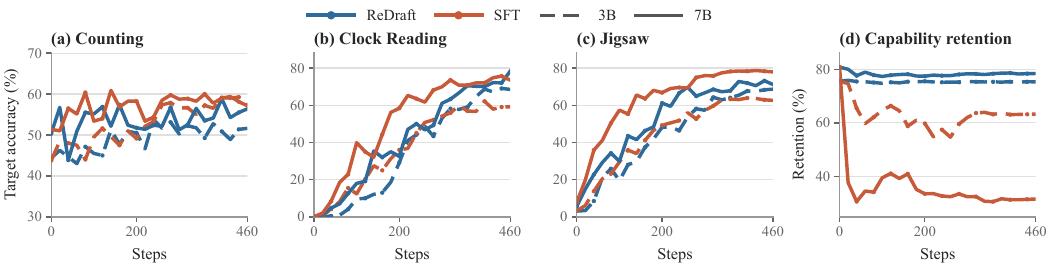}
   \caption{\textbf{Mixed training at both model scales.} All three target
  corpora are interleaved in a single run. Color identifies the objective and  line style the model scale. Panels (a--c) track the three target tasks separately; (d) shows original-task retention over $8$ benchmarks.}
  \label{fig:mixed-training}
\end{figure}

\paragraph{Mixed training.} Interleaving all three target corpora tests
whether \method{}'s retention advantage survives without task boundaries
(Figure~\ref{fig:mixed-training}). At the final checkpoint on 3B, \method{}
exceeds SFT in mean target accuracy ($62.9$ versus $59.7$) while retaining
$75.3$ versus $63.3$ on the original tasks ($0.2$ versus $12.3$ points of
forgetting). On 7B it acquires the tasks more slowly and ends just short of
SFT ($68.6$ versus $69.5$), while retaining $78.4$ versus $31.6$. The per-task panels show where the two means
come apart: \method{} leads on Clock Reading at both scales and on 3B Jigsaw,
while SFT leads on 3B Counting and on 7B Jigsaw. The retention advantage
holds at both scales. 

\paragraph{Sequential Training.}
\label{app:sequential-training}

\begin{figure}[ht!]
  \centering
  \includegraphics[width=\linewidth]{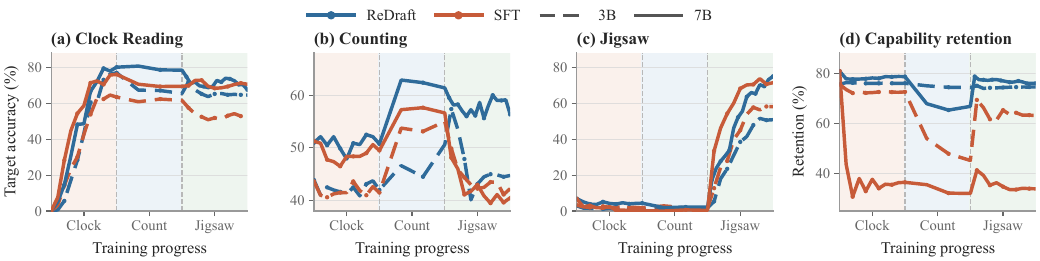}
  \caption{\textbf{Sequential performance across all three target tasks.}
  Panels (a--c) track Clock Reading, Counting, and Jigsaw throughout the
  complete curriculum; (d) shows original-task retention. Color identifies
  the objective and line style the model scale. Backgrounds mark the active
  training stage.}
  \label{fig:sequential-training}
\end{figure}

We further train the three tasks as a Clock$\rightarrow$Counting$\rightarrow$Jigsaw sequential curriculum, where
the same advantage appears: 
\method{} scales to repeated skill acquisition
(Figure~\ref{fig:sequential-training}). At the end of the curriculum on 3B,
\method{} reaches a three-task mean of $53.4$ versus SFT's $51.0$ and retains $74.5$ versus $63.1$ on the original tasks, while SFT leads on the final-stage Jigsaw score ($58.1$ versus $50.9$). On 7B, \method{} ends ahead on Clock Reading and Jigsaw, with a three-task mean of $66.3$ versus $60.9$ and retention of $75.9$ versus $33.9$.

\section{Data- and Parameter-Space Analysis}
\label{sec:ablation-weightspace}

Since \method{} matches SFT on the target task while forgetting far less, and
dominates OPSD on both axes, we ask where that behavior comes from. This
section examines the targets \method{} trains on and the parameter update they
induce. All three objectives start from the same pre-trained weights, so their
updates can be compared directly.

\begin{figure}[!ht]
  \centering
  \includegraphics[width=\linewidth]{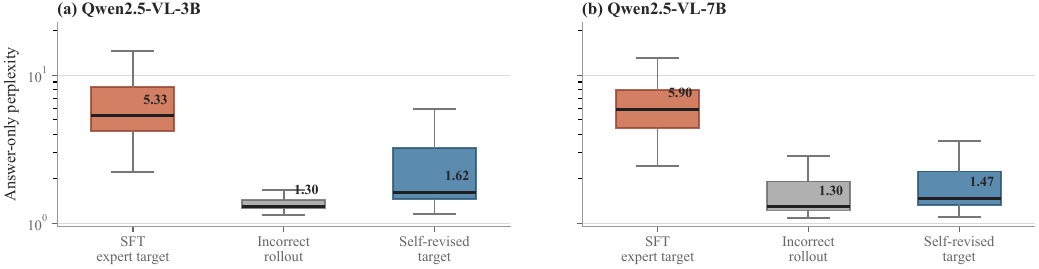}
  \caption{\textbf{Self-revised targets are more policy-proximal than external SFT targets.} Answer-only PPL is computed under the corresponding frozen base model and shown on a logarithmic scale; each box pools $400$ paired examples from each of three tasks.}
  \label{fig:ppl-main}
\end{figure}

\paragraph{Revised targets retain \emph{policy proximity} in the data.}
We first examine how likely the training targets are under the generating
model, using answer-only perplexity (PPL) under the corresponding frozen
base model. For each model and task we score three responses for the same
prompts: the external expert trajectory, the model's own incorrect rollout,
and its verified self-revision, pooling the three tasks in the aggregate
result. Figure~\ref{fig:ppl-main} shows much lower median PPL for
self-revisions than for expert trajectories at both scales ($1.62$ versus
$5.33$ on 3B, $70\%$ lower; $1.47$ versus $5.90$ on 7B, $75\%$ lower). Own
rollouts are lowest ($1.30$ at both scales), providing a reference for the
model's original behavior. Low PPL indicates higher likelihood under that
model, not task correctness. Verified self-revisions have PPL close to this
reference, supporting \emph{policy proximity}, while verification separately
establishes correctness.

\begin{wrapfigure}{r}{0.48\textwidth}
  \vspace{-1\intextsep}
  \centering
  \includegraphics[width=\linewidth]{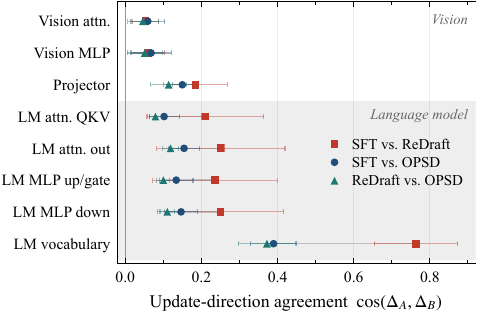}
  \vspace{-1.4\intextsep}
  \caption{\textbf{Agreement between the update directions of different
  methods}, per module group. Markers are means over all pairs.}
  \label{fig:direction}
  \vspace{-1\intextsep}
\end{wrapfigure}

\paragraph{\emph{Explicit target supervision} steers the update in SFT's
direction.} We next examine whether \method{}'s explicit targets induce
updates aligned with SFT's. Training each objective from the base model, we take the cosine $\cos(\Delta_A,\Delta_B)$ between the
updates of two objectives, per module group.
Figure~\ref{fig:direction} localizes the answer: vision blocks
are indistinguishable ($0.05$--$0.06$), language-model linear layers favor
\method{} ($+0.24$ vs. $+0.13$), and the vocabulary tensors show the
largest gap ($+0.76$ vs. $+0.39$). Over all parameters
$\cos(\Delta_{\text{SFT}},\Delta_{\methodm})=+0.272$ against $+0.135$ for
SFT--OPSD. This stronger alignment, especially in the vocabulary tensors,
is consistent with the explicit token targets shared by \method{} and SFT.
Paired results are reported in Appendix~\ref{app:update-direction}.

\begin{figure}[!ht]
  \centering
  \vspace{-0.5\intextsep}
  \includegraphics{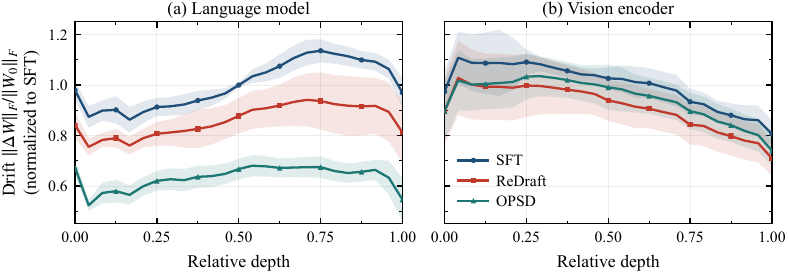}
  \vspace{-0.6\intextsep}
  \caption{\textbf{Update magnitude by network depth.} Each curve is
  normalized by the layer-mean drift so that the two
  model scales can be pooled; curves are means over all task--model pairs.}
  \label{fig:magnitude}
  \vspace{-0.5\intextsep}
\end{figure}

\paragraph{\method{} occupies the intermediate update regime.} We then measure
how far each objective moves the weights. For every tensor we take the
relative displacement $\lVert\Delta W\rVert_F/\lVert W_0\rVert_F$ from its
pre-trained value, aggregate it by module, and average over the three tasks
and both model scales. Figure~\ref{fig:magnitude} shows
the resulting profile by depth: the objectives separate throughout the
language model in the order $\text{SFT}>\methodm>\text{OPSD}$, while the
vision-encoder curves nearly coincide. Thus \method{} changes the weights
less than SFT but more than OPSD. Aggregate displacements are
$(6.52,5.59,4.13)\times10^{-3}$ for SFT, \method{}, and OPSD, respectively
(Appendix~\ref{app:summary}).

\paragraph{The update is concentrated rather than diffuse.} A larger update
need not be a more scattered one. Recent work counts the directions an update
uses by the effective rank of the update matrix, and reports that on-policy
distillation uses fewer of them than RL~\citep{cai2026foresee}. Measured
over the language model's attention and MLP matrices, \method{} is once more
intermediate: its effective rank is $1454.7$, below OPSD's $1506.4$ and above
SFT's $1397.6$. Its larger update is therefore packed into fewer directions
than OPSD's rather than spread more widely. A second concentration measure
gives the same ordering on average; both are defined and reported per setting
in Appendix~\ref{app:spectral-geometry}.

\begin{figure}[ht!]
  \centering
  \includegraphics[width=\linewidth]{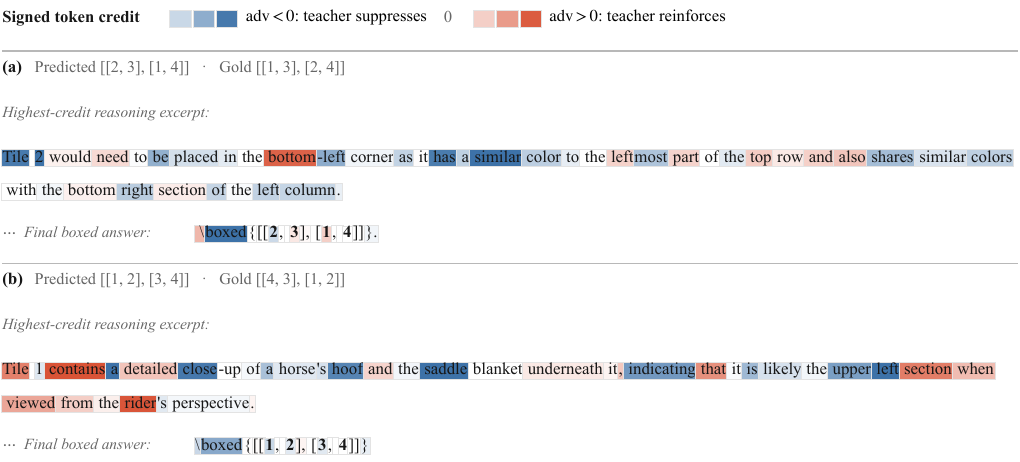}
  \caption{\textbf{OPSD credit peaks away from the task-deciding answer.}
  Token backgrounds show signed sampled-token credit: blue denotes
  $\mathrm{adv}<0$ (teacher suppression), red $\mathrm{adv}>0$ (teacher
  reinforcement), and darker color indicates larger $|\mathrm{adv}|$ under one shared
  scale.}
  \label{fig:token-credit}
\end{figure}

\paragraph{OPSD's token credit misses the decisive tokens.}
Section~\ref{sec:selfrevised} explains why OPSD's token-level supervision
need not focus on task-critical errors. Figure~\ref{fig:token-credit}
examines this issue on two rollouts, colouring each token by
$\mathrm{adv}_n=\log p_T(\hat y_n)-\log p_S(\hat y_n)$, so that blue marks
tokens the teacher suppresses and red ones it reinforces. The answer digits
that alone decide the task carry little of the weight: most of it is spread
over incidental reasoning, style, and formatting tokens, and some answer
tokens are even pushed the wrong way. OPSD can thus know the correction and
still fail to put its signal where the decision is made, whereas \method{}
writes the correction out as a verified target.

\section{Revision Capability}
\label{sec:revision-capability}

\begin{wrapfigure}{r}{0.40\textwidth}
  \vspace{-\intextsep}
  \centering
  \includegraphics[width=\linewidth]{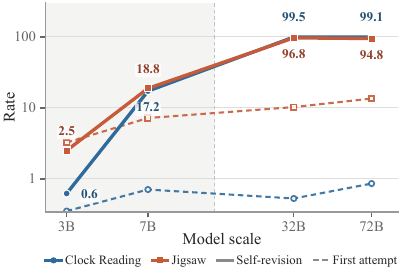}
  \vspace{-1.5\intextsep}
  \caption{\textbf{Revision against model scale under the simplified
  instruction.} \textbf{Solid:} fraction of failed rollouts whose revision the verifier
  accepts. \textbf{Dashed:} fraction of prompts the same model already answers
  correctly on its first attempt. Shading marks the scales at which revision
  mostly fails.}
  \label{fig:revision-scaling}
  \vspace{-\intextsep}
\end{wrapfigure}

\paragraph{Scaling of Self-revision.} \method{} requires the model
to repair its own rollout once the reference is in context
(Assumption~\ref{as:correction}), and how hard that repair is depends on how
much the instruction gives away. The instruction used above states the target
answer, which makes the response easier to repair, a shortcut that STaR~\citep{zelikman2022star} also warns for reasoning unfaithful: the model can carry that
answer over and adjust the reasoning around it. We therefore simplify the
prompt, supplying the expert trajectory alone and asking for the revision
that makes the response correct, without stating what the answer should be
(Appendix~\ref{app:revision-prompt}). Under the simplified prompt the scales
separate sharply (Figure~\ref{fig:revision-scaling}): the verifier accepts
$0.6\%$ and $2.5\%$ of 3B revisions on Clock Reading and Jigsaw, against
$17.2\%$ and $18.8\%$ at 7B and $99.5\%$ and $96.8\%$ at 32B, after which the
rate saturates. At 3B the revision usually returns the original reading
unchanged. Across the same four scales, first-attempt accuracy rises only from
$0.0\%$ at 3B to $0.9\%$ at 72B on Clock Reading, and from $3.3\%$ to $13.5\%$
on Jigsaw, so what scales here is the ability
to revise against a reference, not the ability to solve the task.

\begin{figure}[ht!]
  \centering
  \includegraphics[width=\linewidth]{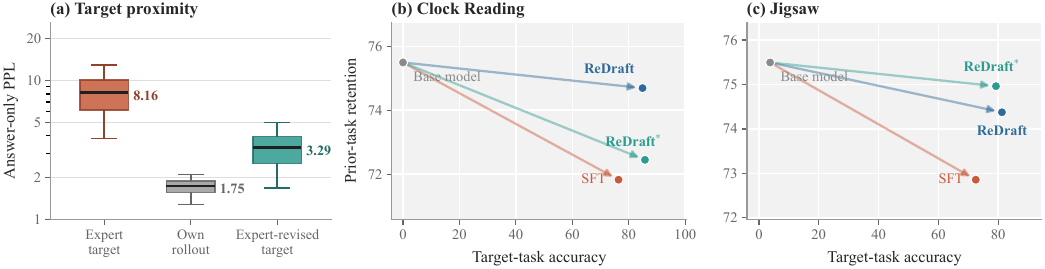}
  \caption{\textbf{Expert revision of a 3B rollout.} (a) Answer-only PPL under
  Qwen2.5-VL-3B, pooled over Clock Reading and Jigsaw. (b--c) Target-task
  accuracy against prior-task retention, with arrows originating at the base
  model; \method{}$^{*}$ denotes the variant whose revisions are produced by
  the external expert rather than by the 3B model itself.}
  \label{fig:expert-revision}
\end{figure}

\paragraph{Leveraging the revision capability of an external expert.} How
should a model that cannot yet revise itself be cold-started? Here we keep the revision step
and move it outside the model: the same expert that supplies $y^{\star}$ is
asked to edit the 3B rollout, changing only what requires correction and
leaving the rest of the trajectory untouched. The expert is used for its
ability to revise rather than as a trajectory to imitate, so as much of the model's own response survives, and we call this variant \method{}$^{*}$.
Figure~\ref{fig:expert-revision}a shows that pooled over the two tasks, their median
answer-only PPL is $3.29$, against $8.16$ for the expert's own trajectory and
$1.75$ for the model's own rollout. An externally revised target is therefore far closer to the
model's distribution than the expert trace it would otherwise imitate, though
not as close as a revision the model writes itself. Training on these targets
recovers most of \method{}'s behavior (Figure~\ref{fig:expert-revision}b,c): on Clock Reading
\method{}$^{*}$ reaches $85.8$ target accuracy against SFT's $76.4$ while
retaining $72.5$ against $71.8$, and on Jigsaw it reaches $79.3$ against
$72.5$ while retaining $75.0$ against $72.9$, the least forgetting of the
three objectives. The 3B model therefore acquires the new task while
retaining more of its prior capability than SFT allows.

\section{Conclusion}
\label{sec:conclusion}

We studied continual post-training on tasks the model cannot yet perform. Such
a model has to be shown what a correct answer looks like, and the
demonstrations that do this lie outside its own distribution, so fitting them
drives forgetting. On-policy self-distillation avoids that shift by staying on
the model's own rollouts, but its correction survives only as a token-level
divergence from a reference-conditioned teacher, which need not fall on the
decisions that settle the task. \method{} makes the correction explicit without
leaving the model's own distribution: an incorrect rollout is edited as little
as the reference allows, kept only if a verifier accepts it, and trained on as
ordinary token-level supervision. Across three tasks, two model scales,
and both mixed and sequential curricula, it recovers SFT's acquisition at a
fraction of its forgetting and improves on OPSD along both axes.

\section*{Limitations}

\method{} rests on the learner's correction competence (Assumption~\ref{as:correction}), which holds comfortably here (Appendix~\ref{app:yield}) but remains empirical. If the learner cannot revise reliably, no self-revised targets can be produced. Nevertheless, we propose \method{}$^{*}$ as a more effective cold-start alternative: a stronger expert revises the learner's rollout rather than providing a trajectory for direct distillation. This mitigates but does not eliminate the assumption, since a task that defeats even expert-assisted repair would still leave no usable
targets. \method{} also requires an automatic verifier, so open-ended generation would need a learned or human proxy whose errors enter the targets directly. Each failed rollout costs an extra generation pass. We construct the revised set once from the policy that begins training, so the targets become staler as the parameters move; online revision should restore proximity but remains untested. Finally, we represent the self-distillation family by OPSD as formulated in Equation~\ref{eq:opd}, without its pointwise clipping, and do not evaluate SDFT separately.



\bibliography{iclr2026_conference}
\bibliographystyle{iclr2026_conference}

\appendix

\section{Derivation of the Idealized Target}
\label{app:derivation}

Section~\ref{sec:selfrevised} states three results without derivation: that the
trust-region problem of Equation~\ref{eq:trust-region} is solved by a tilted
distribution, that this tilt turns into a hard correctness constraint in the
limit of a binary verifier, and that the resulting target costs $1/Z_\theta(x)$
samples to obtain by rejection whereas a revision costs $1/S_\theta(x)$. This
appendix derives the three results and, for each one, says what it is used
for. Notation follows the main text throughout.

\subsection{Setting and Notation}
\label{app:notation}

Everything below concerns one prompt $x$ at one point during training.

\begin{itemize}[leftmargin=1.4em,itemsep=2pt,topsep=2pt]
  \item $y$ is a complete response and $\mathcal{Y}$ is the set of responses
  the model can emit; every sum below runs over $\mathcal{Y}$.
  \item $y^{\star}$ is the expert reference response paired with $x$ in the
  demonstration set $\Dcal$.
  \item $v(y,y^{\star})\in\{0,1\}$ is the verifier, equal to $1$ exactly when
  $y$ solves the task, and $\Ccal(x)=\{y:v(y,y^{\star})=1\}$ is the set of
  responses it accepts. When the verifier is used as a reward we write
  $r(y)=v(y,y^{\star})$.
  \item $\pi_\theta(\cdot\mid x)$ is the current policy.
  \item $\beta>0$ weights the KL penalty. We denote the corresponding
  optimum by $\pi^\star_\beta$ and its $\beta\to0^+$ limit by $\pi^\star$.
  These are idealized targets, not claims about the policy reached by training.
  \item $Z_\theta(x)=\Pr_{y\sim\pi_\theta(\cdot\mid x)}[y\in\Ccal(x)]$
  is the acceptance probability of an initial rollout.
\end{itemize}

\subsection{Solving the Trust-Region Problem}
\label{app:tilted}

Written out, the objective maximized in Equation~\ref{eq:trust-region} is
\begin{equation*}
  J(\pi)
  =\underbrace{\sum_{y}\pi(y\mid x)\,r(y)}_{\text{expected verifier reward}}
  \;-\;\beta\underbrace{\sum_{y}\pi(y\mid x)\log\frac{\pi(y\mid x)}{\pi_\theta(y\mid x)}}
  _{\mathrm{KL}\left(\pi(\cdot\mid x)\,\|\,\pi_\theta(\cdot\mid x)\right)},
\end{equation*}
maximized over distributions $\pi(\cdot\mid x)$ subject to
$\sum_y\pi(y\mid x)=1$ and $\pi(y\mid x)\ge0$. Attaching a multiplier
$\lambda$ to the normalization constraint gives the Lagrangian
$L(\pi,\lambda)=J(\pi)-\lambda\big(\sum_y\pi(y\mid x)-1\big)$, whose
derivative with respect to the probability of one response is
\begin{equation*}
  \frac{\partial L}{\partial\pi(y\mid x)}
  = r(y)-\beta\left[\log\frac{\pi(y\mid x)}{\pi_\theta(y\mid x)}+1\right]-\lambda .
\end{equation*}
Setting it to zero and solving for $\pi(y\mid x)$ gives
\begin{equation*}
  \log\frac{\pi(y\mid x)}{\pi_\theta(y\mid x)}=\frac{r(y)-\lambda}{\beta}-1
  \qquad\Longrightarrow\qquad
  \pi(y\mid x)=\pi_\theta(y\mid x)\,\exp\!\big(r(y)/\beta\big)\cdot
               e^{-\lambda/\beta-1}.
\end{equation*}
The trailing factor is the same for every response, so it is fixed by
normalization rather than by solving for $\lambda$, and we obtain
\begin{equation}
  \pi^{\star}_{\beta}(y\mid x)
  =\frac{\pi_\theta(y\mid x)\,\exp\!\big(r(y)/\beta\big)}{N_\beta(x)},
  \qquad
  N_\beta(x)=\sum_{y'}\pi_\theta(y'\mid x)\exp\!\big(r(y')/\beta\big).
  \label{eq:app-tilted}
\end{equation}

\subsection{The Limit of a Binary Verifier}
\label{app:limit}

We now specialize Equation~\ref{eq:app-tilted} to our reward. Because
$r(y)=v(y,y^{\star})$ takes only the values $0$ and $1$, the factor
$\exp(r(y)/\beta)$ takes only the value $e^{1/\beta}$ on accepted responses
and $1$ on rejected ones. The normalizer therefore splits into two sums,
\begin{equation*}
  N_\beta(x)
  =\sum_{y\in\Ccal(x)}\pi_\theta(y\mid x)\,e^{1/\beta}
   +\sum_{y\notin\Ccal(x)}\pi_\theta(y\mid x)
  = Z_\theta(x)\,e^{1/\beta}+\big(1-Z_\theta(x)\big),
\end{equation*}
and substituting it back into Equation~\ref{eq:app-tilted} gives
\begin{equation}
  \pi^{\star}_{\beta}(y\mid x)=
  \begin{cases}
    \dfrac{\pi_\theta(y\mid x)}{Z_\theta(x)+\big(1-Z_\theta(x)\big)e^{-1/\beta}}, & y\in\Ccal(x),\\[2.4ex]
    \dfrac{\pi_\theta(y\mid x)}{Z_\theta(x)\,e^{1/\beta}+\big(1-Z_\theta(x)\big)}, & y\notin\Ccal(x),
  \end{cases}
  \label{eq:app-cases}
\end{equation}
where the first line follows after dividing numerator and denominator by
$e^{1/\beta}$.

Before taking a limit it helps to see what $\beta$ actually controls here. In
the objective $\beta$ is the weight of the proximity penalty, but in the
solution it enters only through the factor $e^{1/\beta}$, which is the
advantage that being accepted confers. Comparing an accepted $y$ with a
rejected $y'$, and summing the second line of
Equation~\ref{eq:app-cases} over all rejected responses,
\begin{equation*}
  \frac{\pi^{\star}_{\beta}(y\mid x)}{\pi^{\star}_{\beta}(y'\mid x)}
  =\frac{\pi_\theta(y\mid x)}{\pi_\theta(y'\mid x)}\cdot e^{1/\beta},
  \qquad
  \sum_{y\notin\Ccal(x)}\pi^{\star}_{\beta}(y\mid x)
  =\frac{1-Z_\theta(x)}{Z_\theta(x)\,e^{1/\beta}+1-Z_\theta(x)} .
\end{equation*}
A large $\beta$ makes $e^{1/\beta}\!\approx\!1$, so the optimum stays close to
$\pi_\theta(\cdot\mid x)$ and leaves nearly the original mass $1-Z_\theta(x)$ on
responses the verifier rejects; shrinking $\beta$ raises $e^{1/\beta}$ and
drives that mass towards zero. Correctness is thus only ever a soft
preference: at any finite $\beta$ the optimum is a mixture that still places
probability on incorrect responses.


Taking $\beta\to0^{+}$ makes $1/\beta\to\infty$. In the first line of
Equation~\ref{eq:app-cases} the term $e^{-1/\beta}$ vanishes and the
denominator tends to $Z_\theta(x)$, while in the second line $e^{1/\beta}$ diverges
and the whole expression tends to zero. Provided $Z_\theta(x)>0$,
\begin{equation}
  \pi^{\star}(y\mid x)
  =\lim_{\beta\to0^{+}}\pi^{\star}_{\beta}(y\mid x)
  =\frac{\pi_\theta(y\mid x)\,\mathbf{1}\{y\in\Ccal(x)\}}{Z_\theta(x)},
  \label{eq:app-limit}
\end{equation}
which is Equation~\ref{eq:ideal-target}.

Equation~\ref{eq:app-limit} is the conditional law of
$y\sim\pi_\theta(\cdot\mid x)$ given $y\in\Ccal(x)$: correct responses keep
the relative probabilities the current policy gave them and incorrect ones are
dropped. Proximity therefore survives the limit, since the target is not
uniform on $\Ccal(x)$ but still weighted by $\pi_\theta(y\mid x)$, which is

\subsection{Where Each Objective Sits}
\label{app:cost}

Equation~\ref{eq:app-limit} has exactly two ingredients: a proposal,
$\pi_\theta(\cdot\mid x)$, and a verified indicator $\mathbf{1}\{y\in\Ccal(x)\}$
normalized by $Z_\theta(x)$. Each objective of Section~\ref{sec:twoways} changes one
of them, and reading off which one explains both the failures and the fix.

Rejection sampling and RLVR change neither. They draw
$\hat y_1,\hat y_2,\dots$ from $\pi_\theta(\cdot\mid x)$ and keep the first
accepted draw, which is an unbiased sample from
Equation~\ref{eq:app-limit}. The index $N$ of that draw is geometric with
success probability $Z_\theta(x)$, so
\begin{equation*}
  \mathbb{E}[N]
  =\sum_{k\ge1}k\,\big(1-Z_\theta(x)\big)^{k-1}Z_\theta(x)
  =\frac{1}{Z_\theta(x)} ,
\end{equation*}
and no target is ever produced when $Z_\theta(x)=0$, the limit the cold-start
tasks approach with $Z_\theta$ below $1\%$ (Appendix~\ref{app:yield}).

SFT keeps the indicator and discards the $\pi_\theta$ factor. 

We identify the distribution that SFT actually fits. Write
$\delta_{y^{\star}}$ for the point mass at the reference, that is
$\delta_{y^{\star}}(y)=1$ for $y=y^{\star}$ and $0$ for every other response,
%
which is the objective of Equation~\ref{eq:sft}. The target SFT fits is
therefore $\delta_{y^{\star}}$.

We compare $\delta_{y^{\star}}$ with Equation~\ref{eq:app-limit}. Both
put all their mass inside $\Ccal(x)$, so both satisfy the indicator. They
differ in how that mass is distributed: $\pi^{\star}$ spreads it over every
accepted response in proportion to $\pi_\theta(y\mid x)$, whereas
$\delta_{y^{\star}}$ puts all of it on one accepted response chosen without
consulting $\pi_\theta$ at all. This is the precise sense in which SFT
replaces $\pi^{\star}$ by $y^{\star}$: the indicator is kept, the $\pi_\theta$ factor is dropped.
The disagreement is quantitative as well. Equation~\ref{eq:app-limit} assigns
the reference probability
$\pi^{\star}(y^{\star}\mid x)=\pi_\theta(y^{\star}\mid x)/Z_\theta(x)$ while
$\delta_{y^{\star}}$ assigns it $1$, so SFT over-weights the reference by
\begin{equation*}
  \frac{1}{\pi^{\star}(y^{\star}\mid x)}
  =\frac{Z_\theta(x)}{\pi_\theta(y^{\star}\mid x)}\;\ge\;1 ,
\end{equation*}
since $Z_\theta(x)=\sum_{y\in\Ccal(x)}\pi_\theta(y\mid x)$ includes the term
$\pi_\theta(y^{\star}\mid x)$; the two coincide only when the current policy
gives no probability to any other accepted response. The factor grows without
bound as $y^{\star}$ is not sampled from $\pi_\theta(\cdot \mid x)$, which is the cold-start regime,
and one quantity therefore drives both of SFT's observed behaviors: an
explicit target that works from near-zero accuracy, and an update large enough
to forget.

Self-distillation keeps the proposal and replaces the indicator. It targets the
same $\pi^{\star}$, but does not use the verifier during training, so it cannot
use the characterization in Equation~\ref{eq:app-limit} and instead puts the
reference-conditioned model in its place; dividing that model by
$\pi_\theta(y\mid x)$ then writes it as a tilt of the same form as
Equation~\ref{eq:app-tilted},
\begin{equation}
  \pi^{\star}(y\mid x)
  \;\approx\;\pi_\theta(y\mid x,y^{\star})
  \;=\;\pi_\theta(y\mid x)\,\exp\!\big(\Delta_\theta(y)\big),
  \qquad
  \Delta_\theta(y)
  =\log\frac{\pi_\theta(y\mid x,y^{\star})}{\pi_\theta(y\mid x)},
  \label{eq:app-opsd-tilt}
\end{equation}
Comparing Equation~\ref{eq:app-opsd-tilt} with
Equation~\ref{eq:app-limit}, the hard indicator has become the soft exponent
$\Delta_\theta$, which no verifier constrains. Two consequences follow, and
they are the two defects the main text reports: the target need not place all
its mass on $\Ccal(x)$, so correctness is not guaranteed, and the exponent is
carried by per-token terms $a_n$ that need not sit at the positions deciding
acceptance, which is what $\eta$ measures.

\method{} keeps the indicator and replaces the proposal. A candidate is
produced in two steps: a rollout $\hat y\sim\pi_\theta(\cdot\mid x)$, followed
by a revision $\tilde y\sim R_\theta(\cdot\mid x,y^{\star},\hat y)$, where
$R_\theta$ is the distribution of the response the model emits when given the
prompt, the reference and its own rollout, and returns $\hat y$ unchanged when
that rollout is already accepted. Only the candidate is used for training, so
what we need is the law of $\tilde y$ on its own, obtained by averaging over
the intermediate rollout:
\begin{equation}
  \rho_\theta(y\mid x)
  =\sum_{\hat y}\pi_\theta(\hat y\mid x)\,
    R_\theta\!\left(y\,\middle|\,x,y^{\star},\hat y\right).
  \label{eq:app-proposal}
\end{equation}
Here $\rho_\theta(\cdot\mid x)$ is the distribution of responses produced by
``draft an answer, then edit it against the reference'', in the same sense
that $\pi_\theta(\cdot\mid x)$ is the distribution produced by ``answer
directly''. It takes the place of $\pi_\theta(\cdot\mid x)$ as the proposal,
and nothing else about the procedure changes.

The test itself is unchanged: $\Ccal(x)$ is the same set of verifier-accepted
responses as before, independent of how a candidate was produced, so a
candidate is retained if and only if $\tilde y\in\Ccal(x)$. Under the new
proposal that happens with probability
\begin{equation*}
  S_\theta(x)
  =\Pr_{\tilde y\sim\rho_\theta(\cdot\mid x)}\!\left[\tilde y\in\Ccal(x)\right]
  =\sum_{y\in\Ccal(x)}\rho_\theta(y\mid x) ,
\end{equation*}
the same sum that defines $Z_\theta(x)$ over the same set, but taken against
$\rho_\theta$ instead of $\pi_\theta$.

A training target is the first candidate to pass the test, so its distribution
follows from summing over the number of candidates rejected before it.
Candidates are drawn independently, so the $k$-th of them is the first to be
retained and equals $y$ when the first $k-1$ draws are rejected, of
probability $\big(1-S_\theta(x)\big)^{k-1}$, and the $k$-th draw is $y$ and
accepted, of probability
$\rho_\theta(y\mid x)\mathbf{1}\{y\in\Ccal(x)\}$. Summing over $k$ with
$\sum_{k\ge1}(1-p)^{k-1}=1/p$,
\begin{equation}
  \pi_{\methodm}(y\mid x)
  =\sum_{k\ge1}\big(1-S_\theta(x)\big)^{k-1}
     \rho_\theta(y\mid x)\,\mathbf{1}\{y\in\Ccal(x)\}
  =\frac{\rho_\theta(y\mid x)\,\mathbf{1}\{y\in\Ccal(x)\}}{S_\theta(x)} .
  \label{eq:app-redraft-target}
\end{equation}
Discarding the rejected candidates has therefore sampled $\rho_\theta$
conditioned on acceptance: Equation~\ref{eq:app-redraft-target} is
Equation~\ref{eq:app-limit} with $(\pi_\theta,Z)$ replaced by
$(\rho_\theta,S_\theta)$, derived rather than assumed. The indicator survives,
so every retained target is correct, and the expected number of attempts per
target is $1/S_\theta(x)$.

The two acceptance rates are related exactly, which is where the cold-start
gain comes from. Splitting on whether the rollout already passes, and using
that $R_\theta$ returns $\hat y$ unchanged in that case, so that
$\Pr[\tilde y\in\Ccal(x)\mid\hat y\in\Ccal(x)]=1$,
\begin{equation*}
  S_\theta(x)
  =\underbrace{\Pr\big[\hat y\in\Ccal(x)\big]}_{Z_\theta(x)}\cdot 1
   +\underbrace{\Pr\big[\hat y\notin\Ccal(x)\big]}_{1-Z_\theta(x)}\cdot
    \underbrace{\Pr\big[\tilde y\in\Ccal(x)\,\big|\,\hat y\notin\Ccal(x)\big]}_{r_\theta(x)},
\end{equation*}
which is the identity in Equation~\ref{eq:redraft-target}. Revision keeps the $Z_\theta(x)$ that
rejection sampling already collects and adds the term
$\big(1-Z_\theta(x)\big)r_\theta(x)$ recovered from the failures that
rejection sampling discards. At cold start the first term is negligible and
the second is the entire yield, so $S_\theta(x)>0$ is possible exactly where
$Z_\theta(x)=0$: the events differ because $\tilde y$ is drawn with the
reference in context, from a distribution whose support is not that of
$\pi_\theta(\cdot\mid x)$.

The substitution is not free: $\rho_\theta\neq\pi_\theta$, so
Equation~\ref{eq:app-redraft-target} is a biased surrogate for
Equation~\ref{eq:app-limit}. The minimal-edit instruction biases $\tilde y$
towards $\hat y$ but does not constrain it, and a repair is sometimes a full
rewrite, so we do not bound the discrepancy analytically. What holds is that
$\tilde y$ is emitted by the policy itself, and
Section~\ref{sec:ablation-weightspace} measures the discrepancy that remains
as target perplexity under the generating model.

\section{Per-Task Acquisition and Retention Results}
\label{app:task-level-results}

\subsection{Sampling and Repair Rates}
\label{app:yield}

Assumption~\ref{as:correction} requires $r_\theta(x)>0$, and
Equation~\ref{eq:redraft-target} makes \method{}'s supply of training targets
depend on it. We measure both rates on the training corpora by sampling one
rollout per prompt and, when the verifier rejects it, revising it once with
the reference in context (Table~\ref{tab:yield}).

The cold-start tasks show the gap the identity predicts. At 3B the model
solves $0.21\%$ of Jigsaw and $0.03\%$ of Clock Reading prompts unaided, so
rejection sampling collects almost nothing, but $92.25\%$ and $56.84\%$ of the
failed rollouts are repairable, giving targets for $92.27\%$ and $56.85\%$ of
prompts. On
Counting, where sampling already works, revision still raises the yield from
$65.85\%$ to $92.00\%$.

\begin{table}[!ht]
  \centering
  \small
  \setlength{\tabcolsep}{7pt}
   \caption{\textbf{Sampling success, repair rate, and target yield.} $N$ is the number
  of prompts in the corpus, $Z_\theta$ the fraction solved by the first
  rollout, $r_\theta$ the fraction of the remaining prompts whose revision the
  verifier accepts, and $S_\theta=Z_\theta+(1-Z_\theta)r_\theta$ the fraction
  that contributes a training target.}
  \label{tab:yield}
  \begin{tabular}{@{}llrrrr@{}}
    \toprule
    Task & Model & $N$ & Sampling ($Z_\theta$) & Repair ($r_\theta$) & Yield ($S_\theta$) \\
    \midrule
    Counting      & 3B & 2,000  & $65.85$ & $76.57$ & $92.0$ \\
    Jigsaw        & 3B & 10,000 & $0.21$  & $92.25$ & $92.27$ \\
    Clock Reading & 3B & 10,000 & $0.03$  & $56.84$ & $56.85$ \\
    \midrule
    Counting      & 7B & 2,000  & $77.40$ & $98.23$ & $99.6$ \\
    Jigsaw        & 7B & 10,000 & $6.09$  & $94.0$ & $94.37$ \\
    Clock Reading & 7B & 10,000 & $0.67$  & $64.97$ & $65.2$ \\
    \bottomrule
  \end{tabular}
\end{table}

\subsection{RLVR at Cold Start}
\label{app:rlvr-coldstart}

Section~\ref{sec:selfrevised} attributes RLVR's weakness at cold start to the
absence of positive rollouts rather than to insufficient training. We test this
by running GRPO on the two cold-start tasks at both scales, sampling $G=8$
rollouts for each of $32$ prompts per step at a learning rate of
$2\times10^{-6}$ and dropping groups whose rollouts all receive the same reward.

\begin{figure}[ht!]
  \centering
  \includegraphics[width=\linewidth]{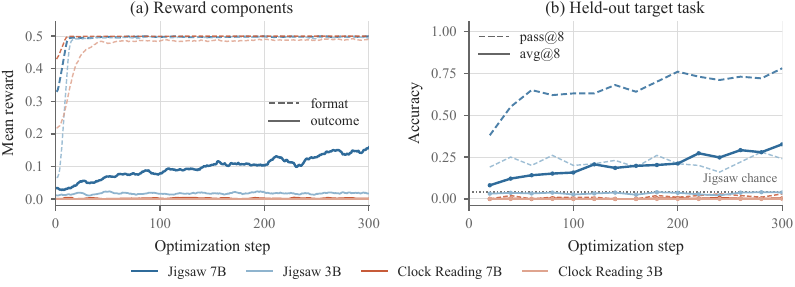}
  \caption{\textbf{GRPO on the two cold-start tasks.} (a) The two reward
  components during training. The format term reaches it in every setting, while the outcome term rises only for Jigsaw at 7B. (b) Accuracy on $100$
  held-out prompts, evaluated with $8$ samples per prompt. The
  dotted line marks the accuracy of guessing a $2\times2$ ordering at random.}
  \label{fig:rlvr-coldstart}
\end{figure}

Figure~\ref{fig:rlvr-coldstart} separates a rising reward from an acquired
skill: the format term saturates everywhere while the outcome term moves only in a single setting.
Clock Reading never leaves zero at either scale, and Jigsaw at 3B sits at the
accuracy of guessing an ordering at random, so neither is learning. Only Jigsaw
at 7B improves, and two properties of that setting supply what the others lack:
its answer space is small enough that a correct ordering is occasionally sampled
by chance, and the 7B model already solves $6.09\%$ of these prompts unaided
against $0.21\%$ at 3B (Table~\ref{tab:yield}). Verifiable rewards therefore
presuppose some of the competence they are meant to build, whereas revision
reaches a $56.85\%$ yield on the same 3B Clock Reading corpus on which GRPO
obtains nothing.

\subsection{Per-Setting Results}
\label{app:detailed-results}

Table~\ref{tab:detailed-results} expands the aggregate comparison of
Section~\ref{sec:main-results} into individual runs. Each target task is
trained separately on Qwen2.5-VL-3B and 7B, and every row is one
task at its selected checkpoint: Target and Prior are
its target-task and prior-capability scores, and Gain and Loss are the
corresponding changes from the base model. Reject Sampling is evaluated only
on Counting, where the base policy already has substantial competence. The means over these rows are the values reported in the
main text and plotted in Figure~\ref{fig:main-results}. The per-setting
entries also show how unevenly forgetting is distributed across runs: SFT
retains $18.92$ on prior tasks for 7B Counting, whereas \method{} retains
$76.00$ on the same run.

\begin{table}[!ht]
  \centering
  \scriptsize
  \setlength{\tabcolsep}{3.8pt}
  \caption{\textbf{Per-setting results.}
  Target and Prior are its scores, while Gain is target
  improvement and Loss is the signed change in prior-task performance relative
  to the corresponding base model, all in percentage points. \method{} matches or exceeds SFT's target score, sometimes over more steps, and retains more prior-task performance in every setting. Reject Sampling is available only for Counting.}
  \label{tab:detailed-results}
  \begin{tabular}{lllrrrrr}
    \toprule
    Task & Model & Method & Steps & Target & Prior & Gain $\uparrow$ & Loss \\
    \midrule
    Counting & 3B & SFT       & 63  & 54.13 & 68.91 & $+10.13$ & $-6.58$ \\
             &    & \method{} & 104 & 56.75 & 76.00 & $+12.75$ & $+0.50$ \\
             &    & Reject Sampling & 42 & 45.00 & 76.24 & $+1.00$ & $+0.75$ \\
             &    & OPSD      & 63  & 51.38 & 75.16 & $+7.38$ & $-0.33$ \\
    \addlinespace
    Jigsaw & 3B & SFT       & 313 & 72.50 & 72.86 & $+68.75$ & $-2.64$ \\
            &    & \method{} & 612 & 81.25 & 74.37 & $+77.50$ & $-1.12$ \\
            &    & OPSD      & 626 & 6.12  & 75.34 & $+2.37$  & $-0.15$ \\
    \addlinespace
    Clock Reading & 3B & SFT       & 313 & 76.38 & 71.83 & $+76.38$ & $-3.67$ \\
                  &    & \method{} & 356 & 84.88 & 74.69 & $+84.88$ & $-0.80$ \\
                  &    & OPSD      & 626 & 43.13 & 75.02 & $+43.13$ & $-0.48$ \\
    \midrule
    Counting & 7B & SFT       & 126 & 64.38 & 18.92 & $+12.13$ & $-61.87$ \\
             &    & \method{} & 124 & 63.62 & 80.16 & $+11.37$ & $-0.63$ \\
             &    & Reject Sampling & 49 & 53.50 & 79.89 & $+1.25$ & $-0.90$ \\
             &    & OPSD      & 126 & 56.63 & 60.06 & $+4.38$ & $-20.73$ \\
    \addlinespace
    Jigsaw & 7B & SFT       & 313 & 75.50 & 73.46 & $+70.25$ & $-7.34$ \\
            &    & \method{} & 596 & 76.88 & 78.75 & $+71.63$ & $-2.04$ \\
            &    & OPSD      & 626 & 19.50 & 78.24 & $+14.25$ & $-2.56$ \\
    \addlinespace
    Clock Reading & 7B & SFT       & 313 & 80.00 & 63.11 & $+79.87$ & $-17.68$ \\
                  &    & \method{} & 408 & 83.63 & 76.02 & $+83.50$ & $-4.77$ \\
                  &    & OPSD      & 626 & 44.50 & 67.67 & $+44.37$ & $-13.12$ \\
    \bottomrule
  \end{tabular}
\end{table}

\subsection{Extended OPSD Training}
\label{app:opsd-extension}

Table~\ref{tab:opsd-detail} breaks Table~\ref{tab:opsd-extension} down by task
and model scale. The longer budget leaves acquisition essentially unchanged in
most cells: on 3B, the gain moves by $1.6$ points on Jigsaw and $0.5$ on Clock
Reading even though the number of updates grows by $50$ to $100\%$, and
Counting stays below $9$ points at either scale. The single substantial
improvement is 7B Clock Reading, which rises from $+44.4$ to $+61.4$ and
accounts for most of the difference between the two budgets. Even there OPSD
ends $22.1$ points below the $+83.5$ that \method{} reaches on the same run.

\begin{table}[!ht]
  \centering
  \small
  \setlength{\tabcolsep}{8pt}
  \caption{\textbf{OPSD under a longer step budget, per task and model scale.} 
Steps is cumulative, and Gain is target-accuracy
  improvement in percentage points over base accuracies of
  $44.00/3.75/0.00$ on 3B and $52.25/5.25/0.13$ on 7B for Counting, Jigsaw,
  and Clock Reading. Averaging the two scales column-wise reproduces
  Table~\ref{tab:opsd-extension}.}
  \label{tab:opsd-detail}
  \begin{tabular}{@{}l rr rr rr rr@{}}
    \toprule
    & \multicolumn{4}{c}{Qwen2.5-VL-3B} & \multicolumn{4}{c}{Qwen2.5-VL-7B} \\
    \cmidrule(lr){2-5}\cmidrule(lr){6-9}
    & \multicolumn{2}{c}{standard} & \multicolumn{2}{c}{extended}
    & \multicolumn{2}{c}{standard} & \multicolumn{2}{c}{extended} \\
    \cmidrule(lr){2-3}\cmidrule(lr){4-5}\cmidrule(lr){6-7}\cmidrule(lr){8-9}
    Task & Steps & Gain & Steps & Gain & Steps & Gain & Steps & Gain \\
    \midrule
    Counting      & 63  & $+7.38$  & 252  & $+8.12$  & 126 & $+4.38$  & 252 & $+8.13$  \\
    Jigsaw        & 626 & $+2.37$  & 939  & $+4.00$  & 626 & $+14.25$ & 939 & $+15.12$ \\
    Clock Reading & 626 & $+43.13$ & 1252 & $+43.63$ & 626 & $+44.37$ & 939 & $+61.37$ \\
    \midrule
    Mean          & 438 & $+17.63$ & 814  & $+18.58$ & 459 & $+21.00$ & 710 & $+28.21$ \\
    \bottomrule
  \end{tabular}
\end{table}
\FloatBarrier

\section{Per-Task Answer-Perplexity Distributions}
\label{app:ppl-detail}

\begin{figure}[ht!]
  \centering
  \includegraphics[width=\linewidth]{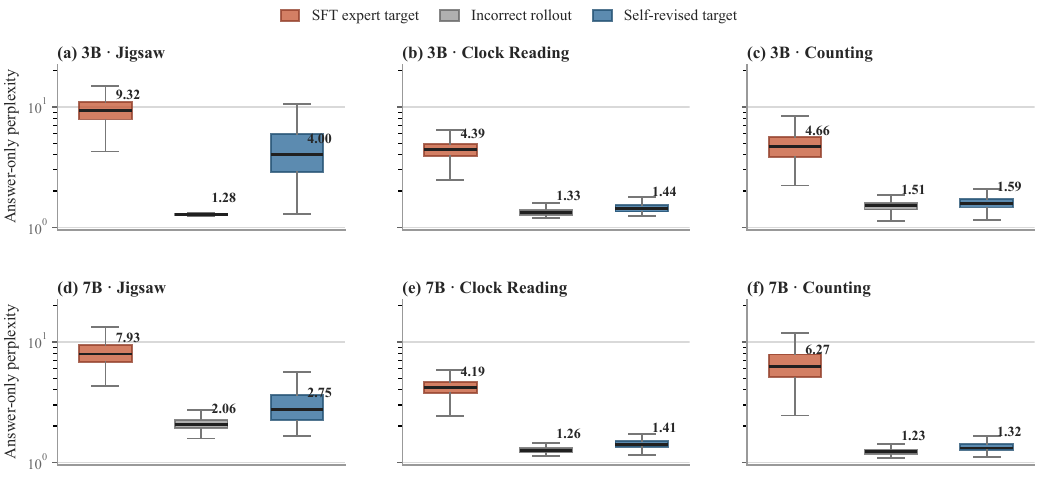}
  \caption{\textbf{Per-task answer-only PPL distributions.} Rows correspond to Qwen2.5-VL-3B/7B and columns to Jigsaw, Clock Reading, and Counting. Each panel contains 400 paired prompts per response source. Boxes span the interquartile range, center lines and labels denote medians, and whiskers extend to $1.5\times$ IQR.}
  \label{fig:ppl-per-task}
\end{figure}

Figure~\ref{fig:ppl-per-task} separates the pooled result by task and model scale. Revision lowers the median in every comparison, and on Counting and Clock Reading it brings the target down to the model's own rollout at both scales: the revised medians are $1.59$ and $1.44$ on 3B and $1.32$ and $1.41$ on 7B, against own-rollout values of $1.51$, $1.33$, $1.23$, and $1.26$, starting from expert targets between $4.19$ and $6.27$. 
Jigsaw is the harder case. At 7B the revised median falls from $7.93$ to $2.75$, close to the $2.06$ of the model's own rollout, but at 3B it stops at $4.00$ against an own-rollout $1.28$, so the revision there stays some way from the model's distribution even though it is verified correct. 
The consistently low PPL of incorrect rollouts again separates policy proximity from correctness: the advantage of self-revision is not merely low PPL, but a verified-correct target that remains closer to the model distribution than the SFT trace.

\section{Per-Setting Weight-Space Results}
\label{app:weightspace-details}

\subsection{Weight-Space Summary}
\label{app:summary}

Table~\ref{tab:summary} collects the weight-space measurements of
Section~\ref{sec:ablation-weightspace}. Three properties of the
update are measured. Its magnitude, the relative displacement of each module
from its pre-trained value, separates the objectives as
$\text{SFT}>\methodm>\text{OPSD}$ throughout the language model and most
sharply at the output head, while the vision encoder and the projector barely
differ between objectives (Appendix~\ref{app:update-magnitude}). Its
concentration, the number of singular directions that carry the update, again
places \method{} between the two baselines
(Appendix~\ref{app:spectral-geometry}). Its direction, the cosine against the
corresponding SFT update, is closer to SFT for \method{} than for OPSD, most
clearly in the vocabulary tensors (Appendix~\ref{app:update-direction}).

\begin{table}[ht!]
  \centering
  \footnotesize
  \renewcommand{\arraystretch}{0.96}
  \caption{\textbf{Weight-space signature and task outcomes.} Weight statistics are means over all task--model pairs; displacements are reported in units of $10^{-3}$ and cosines against the corresponding SFT update. The concentration measures are defined in Appendix~\ref{app:spectral-geometry}. Outcomes follow the Table~\ref{tab:detailed-results}.} 
  \label{tab:summary}
  \begin{tabular}{lccc}
    \toprule
    & SFT & \method{} (ours) & OPSD \\
    \midrule
    \multicolumn{4}{l}{\emph{Update magnitude} ($\times 10^{-3}$)} \\
    \quad All parameters              & 6.52  & 5.59  & 4.13  \\
    \quad Vision encoder              & 9.07  & 8.20  & 8.47  \\
    \quad Projector                   & 7.90  & 6.90  & 7.00  \\
    \quad LM linear layers            & 7.41  & 6.43  & 4.60  \\
    \quad Output head                 & 14.27 & 10.71 & 3.70  \\
    \addlinespace
    \multicolumn{4}{l}{\emph{Spectral geometry}} \\
    \quad Aggregate effective rank    & 1397.6 & 1454.7 & 1506.4 \\
    \quad Top-$1\%$ norm share (\%)   & 59.0 & 54.6 & 53.7 \\
    \quad Leading-direction share $\sigma_1/\lVert\Delta W\rVert_F$ & 0.289 & 0.258 & 0.251 \\
    \addlinespace
    \multicolumn{4}{l}{\emph{Update direction}, $\cos(\Delta,\Delta_{\text{SFT}})$} \\
    \quad All parameters              & --- & $+0.272$ & $+0.135$ \\
    \quad LM linear layers            & --- & $+0.237$ & $+0.134$ \\
    \quad LM vocabulary               & --- & $\mathbf{+0.764}$ & $+0.390$ \\
    \midrule
    \multicolumn{4}{l}{\emph{Task outcomes}} \\
    \quad Target-task gain $\uparrow$ & $+52.9$ & $\mathbf{+56.9}$ & $+19.3$ \\
    \quad Original-task loss $\downarrow$ & $-16.6$ & $\mathbf{-1.5}$ & $-6.2$ \\
    \bottomrule
  \end{tabular}
\end{table}

\subsection{Update Magnitude}
\label{app:update-magnitude}

Table~\ref{tab:magnitude-detail} breaks the displacements averaged in
Table~\ref{tab:summary} down by task and model scale, one row per objective
and module group. The aggregate ordering of Figure~\ref{fig:magnitude} holds
separately in each of them: total displacement, LM-linear drift, and
output-head drift always satisfy $\text{SFT}>\methodm>\text{OPSD}$. The gap is
smallest on Jigsaw, where \method{} nearly matches SFT, and larger on Counting
and Clock Reading.

\begin{table}[ht!]
  \centering
  \scriptsize
  \setlength{\tabcolsep}{4pt}
  \caption{\textbf{Detailed update magnitude}
  ($\times 10^{-3}$). ``All'' covers all measured parameters; vision encoder,
  projector, and LM-linear values are energy-aggregated over matrix weights,
  while output head denotes \texttt{lm\_head.weight}. The task--model pairs
  listed here are the observations averaged in Table~\ref{tab:summary}.}
  \label{tab:magnitude-detail}
  \begin{tabular}{lllrrrrr}
    \toprule
    Task & Model & Method & All & Vision enc. & Projector & LM linear & Output head \\
    \midrule
    Counting & 3B & SFT       & 3.862 & 5.794 & 4.895 & 4.138 & 3.945 \\
             &    & \method{} & 3.261 & 5.352 & 4.184 & 3.440 & 2.910 \\
             &    & OPSD      & 2.856 & 5.621 & 4.053 & 2.860 & 1.564 \\
    \addlinespace
    Clock Reading & 3B & SFT       & 6.882 & 10.464 & 8.676 & 7.063 & 9.876 \\
                  &    & \method{} & 5.050 & 7.477  & 6.760 & 5.396 & 5.486 \\
                  &    & OPSD      & 4.511 & 8.950  & 8.950 & 4.421 & 2.897 \\
    \addlinespace
    Jigsaw & 3B & SFT       & 8.257 & 11.624 & 9.756 & 7.853 & 17.114 \\
            &    & \method{} & 8.033 & 11.058 & 9.149 & 7.629 & 16.984 \\
            &    & OPSD      & 5.215 & 11.097 & 8.093 & 4.893 & 4.369 \\
    \midrule
    Counting & 7B & SFT       & 4.070 & 5.679 & 4.745 & 5.467 & 5.772 \\
             &    & \method{} & 3.365 & 5.510 & 4.457 & 4.440 & 4.408 \\
             &    & OPSD      & 2.869 & 5.928 & 4.436 & 3.658 & 2.475 \\
    \addlinespace
    Clock Reading & 7B & SFT       & 7.293 & 8.764 & 8.761 & 9.160 & 22.151 \\
                  &    & \method{} & 5.836 & 8.528 & 7.358 & 7.507 & 13.653 \\
                  &    & OPSD      & 4.327 & 8.595 & 8.229 & 5.514 & 4.870 \\
    \addlinespace
    Jigsaw & 7B & SFT       & 8.755 & 12.114 & 10.587 & 10.808 & 26.785 \\
            &    & \method{} & 8.003 & 11.265 & 9.481  & 10.169 & 20.834 \\
            &    & OPSD      & 5.004 & 10.633 & 8.255  & 6.267  & 5.995 \\
    \bottomrule
  \end{tabular}
\end{table}

\subsection{Spectral Geometry}
\label{app:spectral-geometry}

The concentration measures are computed from the singular values
$\sigma_1\ge\dots\ge\sigma_r$ of the update $\Delta W=W-W_0$ of each Attention
and MLP weight matrix of the language model, using the full spectrum
($r=\min(m,n)$, no truncation) and averaging over those matrices. Writing
$\bar\sigma_i=\sigma_i/\sum_j\sigma_j$ for the normalized spectrum, the
effective rank is $\exp(-\sum_i\bar\sigma_i\log\bar\sigma_i)$, the exponential
of its entropy, and is lower when the update is carried by fewer directions.
The top-$k$ norm share is
$\big(\sum_{i\le k}\sigma_i^2\big/\sum_j\sigma_j^2\big)^{1/2}$ and is higher
when it is carried by fewer directions; we report it at
$k=\lceil r/100\rceil$, the top $1\%$ of directions, and at $k=1$, where it
reduces to $\sigma_1/\lVert\Delta W\rVert_F$, the fraction of the update
carried by its single leading direction. All three are invariant to rescaling
$\Delta W$, so they describe the shape of the update rather than its size.

Table~\ref{tab:spectral-detail} reports the three measures for each
task. All three place SFT as the most concentrated objective
and \method{} as more concentrated than OPSD.

\begin{table}[ht!]
  \centering
  \small
  \setlength{\tabcolsep}{6pt}
  \caption{\textbf{Detailed spectral geometry.} Norm shares are reported as
  percentages. Effective rank is bounded by the matrix dimensions, which
  differ between scales, so it should be compared between methods within a row
  rather than across scales.}
  \label{tab:spectral-detail}
  \begin{tabular}{lllrrr}
    \toprule
    Task & Model & Method & Effective rank & Top-$1\%$ share (\%) & $\sigma_1/\lVert\Delta W\rVert_F$ \\
    \midrule
    Counting & 3B & SFT       & 1114.0 & 57.2 & 0.299 \\
             &    & \method{} & 1142.6 & 53.6 & 0.273 \\
             &    & OPSD      & 1162.6 & 53.0 & 0.266 \\
    \addlinespace
    Clock Reading & 3B & SFT       & 958.1  & 58.3 & 0.300 \\
                  &    & \method{} & 1040.1 & 53.9 & 0.278 \\
                  &    & OPSD      & 1077.9 & 52.0 & 0.256 \\
    \addlinespace
    Jigsaw & 3B & SFT       & 1032.9 & 53.5 & 0.270 \\
            &    & \method{} & 1043.4 & 52.7 & 0.264 \\
            &    & OPSD      & 1112.3 & 48.5 & 0.235 \\
    \midrule
    Counting & 7B & SFT       & 1879.4 & 68.3 & 0.359 \\
             &    & \method{} & 1951.2 & 57.1 & 0.258 \\
             &    & OPSD      & 1982.1 & 58.5 & 0.277 \\
    \addlinespace
    Clock Reading & 7B & SFT       & 1602.4 & 62.2 & 0.272 \\
                  &    & \method{} & 1682.7 & 60.0 & 0.267 \\
                  &    & OPSD      & 1781.8 & 60.3 & 0.265 \\
    \addlinespace
    Jigsaw & 7B & SFT       & 1799.1 & 54.2 & 0.232 \\
            &    & \method{} & 1868.2 & 50.5 & 0.209 \\
            &    & OPSD      & 1921.9 & 49.7 & 0.205 \\
    \bottomrule
  \end{tabular}
\end{table}

\subsection{Update Direction}
\label{app:update-direction}

Table~\ref{tab:direction-detail} gives the cosines behind
Figure~\ref{fig:direction}, comparing each objective with SFT for every task
and model scale. The full-parameter SFT--\method{} cosine exceeds SFT--OPSD in
every pair, and the vocabulary comparison is likewise unanimous
($0.644$--$0.961$ vs. $0.314$--$0.492$). LM-linear layers favor \method{}
in most settings, with the two Clock Reading settings as exceptions. The
pattern therefore supports a localized claim: \method{} most consistently
recovers SFT's direction where token-level targets enter the embedding and
output geometry.

\begin{table}[ht!]
  \centering
  \scriptsize
  \setlength{\tabcolsep}{6pt}
  \caption{\textbf{Detailed update-direction agreement.} Entries
  are $\cos(\Delta_A,\Delta_B)$ from the same initialization. ``LM linear'' is
  the unweighted mean over attention QKV/output and MLP up--gate/down groups;
  vocabulary denotes \texttt{LLM.embed}.}
  \label{tab:direction-detail}
  \begin{tabular}{lllrrr}
    \toprule
    Task & Model & Pair & All & LM linear & Vocabulary \\
    \midrule
    Counting & 3B & SFT--\method{}       & $+0.224$ & $+0.250$ & $+0.702$ \\
             &    & SFT--OPSD            & $+0.148$ & $+0.182$ & $+0.356$ \\
             &    & \method{}--OPSD      & $+0.079$ & $+0.096$ & $+0.353$ \\
    \addlinespace
    Clock Reading & 3B & SFT--\method{}  & $+0.134$ & $+0.079$ & $+0.644$ \\
                  &    & SFT--OPSD       & $+0.120$ & $+0.095$ & $+0.344$ \\
                  &    & \method{}--OPSD & $+0.098$ & $+0.081$ & $+0.254$ \\
    \addlinespace
    Jigsaw & 3B & SFT--\method{}         & $+0.589$ & $+0.571$ & $+0.961$ \\
            &    & SFT--OPSD             & $+0.121$ & $+0.114$ & $+0.439$ \\
            &    & \method{}--OPSD       & $+0.122$ & $+0.114$ & $+0.448$ \\
    \midrule
    Counting & 7B & SFT--\method{}       & $+0.231$ & $+0.241$ & $+0.660$ \\
             &    & SFT--OPSD            & $+0.182$ & $+0.201$ & $+0.392$ \\
             &    & \method{}--OPSD      & $+0.117$ & $+0.132$ & $+0.400$ \\
    \addlinespace
    Clock Reading & 7B & SFT--\method{}  & $+0.194$ & $+0.102$ & $+0.781$ \\
                  &    & SFT--OPSD       & $+0.141$ & $+0.120$ & $+0.492$ \\
                  &    & \method{}--OPSD & $+0.124$ & $+0.106$ & $+0.470$ \\
    \addlinespace
    Jigsaw & 7B & SFT--\method{}         & $+0.261$ & $+0.177$ & $+0.832$ \\
            &    & SFT--OPSD             & $+0.097$ & $+0.093$ & $+0.314$ \\
            &    & \method{}--OPSD       & $+0.085$ & $+0.084$ & $+0.309$ \\
    \bottomrule
  \end{tabular}
\end{table}

\input{appendices/demo_box_styles.tex}
\clearpage
\section{Case Study}
\label{app:demos}

Below we show one self-revision example for each target task. For every example we
display the visual input, the problem with its ground-truth answer, the
model's incorrect unaided rollout, and the verifier-accepted revised response
used as the \method{} training target.

\subsection{Clock Reading}
\label{app:demo-clock}

\begin{figure}[H]
  \begin{tcolorbox}[demoinput={Input Image}]
    \centering
    \includegraphics[width=0.42\linewidth]{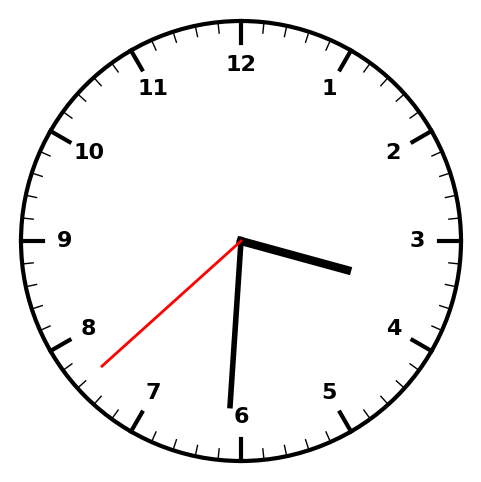}
  \end{tcolorbox}
  \caption{A demonstration of the Clock Reading task.}
  \label{fig:demo-clock}
\end{figure}

Figure~\ref{fig:demo-clock} shows a Clock Reading example.
The unaided rollout misreads all three hands and reports \texttt{06:08:30}.
Given the expert reference only as context for revision, the same model
corrects the reading to the verified answer \texttt{03:30:38}, which is revised
minimally.

\FloatBarrier

\begin{tcolorbox}[demoexample={Example From Clock Reading}]
\textbf{Problem:}\par\vspace{0.25em}
{\footnotesize\DemoMono
You are given an image of an analogue clock. The second hand is red and thin, the minute hand is black and long, and the hour hand is black, shorter and thicker.\par
\vspace{0.35em}
Please answer the following clock-reading question: determine the time displayed on the clock face by locating the hour hand, the minute hand and the second hand, show your detailed step-by-step reasoning process, and wrap the final time in \textbackslash{}boxed\{\} using the zero-padded 12-hour HH:MM:SS format (for example, if the clock shows 3:45:20, end with \textbackslash{}boxed\{03:45:20\}).\par
\vspace{0.35em}
The question is as follows:\par
\vspace{0.35em}
What time is shown on the clock in the given image?\par
\vspace{0.35em}
Now, give your detailed reasoning process in no more than 200 words.\par
}
\vspace{0.6em}
\textbf{Answer:}~\fbox{\texttt{03:30:38}}
\end{tcolorbox}

\begin{tcolorbox}[demowrong]
To determine the time shown on the clock in the given image, we need to analyze the positions of the hour hand, the minute hand, and the second hand.\par
\vspace{0.35em}
1. Hour Hand Analysis:\par
\hspace*{1.5em}- The hour hand is black and longer.\par
\hspace*{1.5em}- It points between the numbers 6 and 7.\par
\hspace*{1.5em}- Counting from 6 to 7, it is approximately halfway between these two numbers.\par
\vspace{0.35em}
2. Minute Hand Analysis:\par
\hspace*{1.5em}- The minute hand is black and longer.\par
\hspace*{1.5em}- It points between the numbers 8 and 9.\par
\hspace*{1.5em}- Counting from 8 to 9, it is approximately halfway between these two numbers.\par
\vspace{0.35em}
3. Second Hand Analysis:\par
\hspace*{1.5em}- The second hand is red and thin.\par
\hspace*{1.5em}- It points between the numbers 10 and 11.\par
\hspace*{1.5em}- Counting from 10 to 11, it is approximately halfway between these two numbers.\par
\vspace{0.35em}
4. Time Calculation:\par
\hspace*{1.5em}- The hour hand is at 6 and 30 seconds (halfway between 6 and 7).\par
\hspace*{1.5em}- The minute hand is at 8 and 30 seconds (halfway between 8 and 9).\par
\hspace*{1.5em}- The second hand is at 10 and 30 seconds (halfway between 10 and 11).\par
\vspace{0.35em}
Therefore, the time shown on the clock is \textbackslash{}boxed\{06:08:30\}.\par
\end{tcolorbox}

\begin{tcolorbox}[demoright]
To determine the time shown on the clock, we need to analyze the positions of the hour hand, the minute hand, and the second hand.\par
\vspace{0.35em}
1. Hour Hand Analysis:\par
\hspace*{1.5em}- The hour hand is black and shorter.\par
\hspace*{1.5em}- It points between the numbers 3 and 4.\par
\hspace*{1.5em}- Counting from 3 to 4, it is approximately halfway between these two numbers.\par
\vspace{0.35em}
2. Minute Hand Analysis:\par
\hspace*{1.5em}- The minute hand is black and longer.\par
\hspace*{1.5em}- It points between the numbers 6 and 7.\par
\hspace*{1.5em}- Counting from 6 to 7, it is approximately halfway between these two numbers.\par
\vspace{0.35em}
3. Second Hand Analysis:\par
\hspace*{1.5em}- The second hand is red and thin.\par
\hspace*{1.5em}- It points between the numbers 7 and 8.\par
\hspace*{1.5em}- Counting from 7 to 8, it is approximately halfway between these two numbers.\par
\vspace{0.35em}
4. Time Calculation:\par
\hspace*{1.5em}- The hour hand is at 3 and 30 seconds (halfway between 3 and 4).\par
\hspace*{1.5em}- The minute hand is at 6 and 30 seconds (halfway between 6 and 7).\par
\hspace*{1.5em}- The second hand is at 7 and 30 seconds (halfway between 7 and 8).\par
\vspace{0.35em}
Therefore, the time shown on the clock is \textbackslash{}boxed\{03:30:38\}.\par
\end{tcolorbox}

\clearpage
\subsection{Counting}
\label{app:demo-count}

\begin{figure}[H]
  \begin{tcolorbox}[demoinput={Input Image}]
    \centering
    \includegraphics[width=0.72\linewidth]{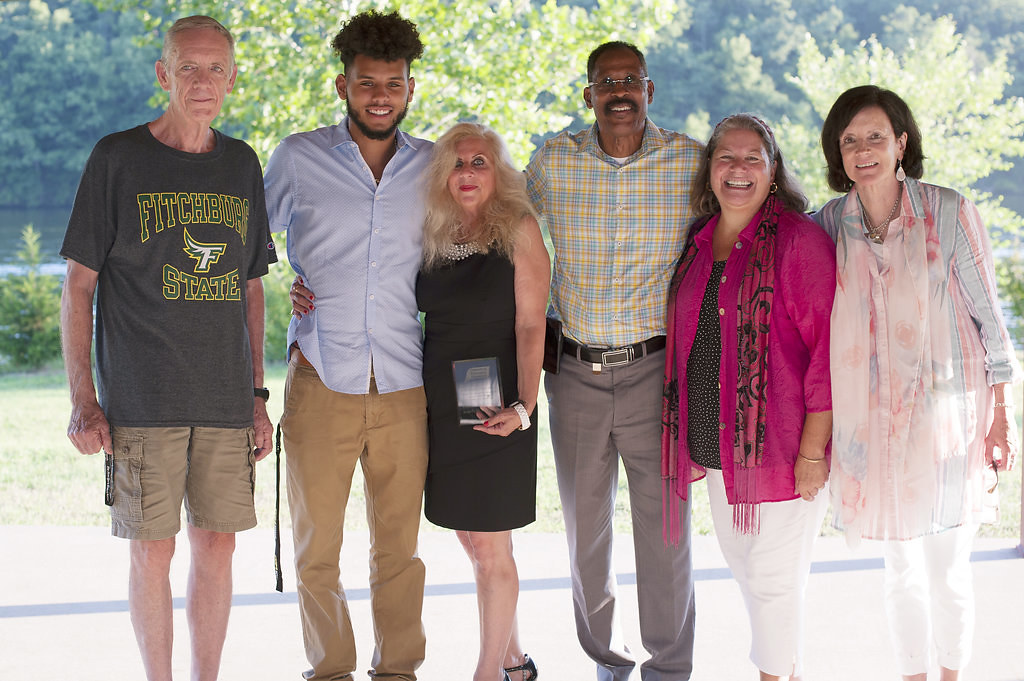}
  \end{tcolorbox}
  \caption{\textbf{A demonstration of the Counting task.}}
  \label{fig:demo-count}
\end{figure}

Figure~\ref{fig:demo-count} shows a Counting example.
The unaided rollout under-counts people in the photograph ($5$ instead of $6$).
After revision minimally against the expert reference, the model adds the missing person
and reports the verified count \texttt{6}, which is revised minimally.

\FloatBarrier

\begin{tcolorbox}[demoexample={Example From Counting}]
\textbf{Problem:}\par\vspace{0.25em}
{\footnotesize\DemoMono
You are given an image. Please answer the following counting question: count how many target objects are in the image, show your detailed step-by-step reasoning process, and wrap the final count in \textbackslash{}boxed\{\}.\par
\vspace{0.35em}
How many people are in this image?\par
}
\vspace{0.6em}
\textbf{Answer:}~\fbox{\texttt{6}}
\end{tcolorbox}

\begin{tcolorbox}[demowrong]
To determine the number of people in the image, I'll follow these steps:\par
\vspace{0.35em}
1. Identify all individuals present in the picture.\par
2. Count each person individually.\par
3. Sum up the total count.\par
\vspace{0.35em}
Step 1: Identifying individuals\par
- There are five people standing together in the image.\par
\vspace{0.35em}
Step 2: Counting individuals\par
- The first person on the left is wearing a gray t-shirt with "FITCHBURG STATE" written on it.\par
- The second person is wearing a light blue button-up shirt and khaki pants.\par
- The third person is holding an award and wearing a black dress.\par
- The fourth person is wearing a yellow plaid shirt with gray pants.\par
- The fifth person is wearing a pink jacket over a black blouse and white pants.\par
\vspace{0.35em}
Step 3: Summing up the total count\par
There are five distinct individuals visible in the image.\par
\vspace{0.35em}
Therefore, the final count is: \textbackslash{}boxed\{5\}\par
\end{tcolorbox}

\begin{tcolorbox}[demoright]
To determine the number of people in the image, I'll follow these steps:\par
\vspace{0.35em}
1. Identify all individuals present in the picture.\par
2. Count each person individually.\par
3. Sum up the total count.\par
\vspace{0.35em}
Step 1: Identifying individuals\par
- There are six people standing together in the image.\par
\vspace{0.35em}
Step 2: Counting individuals\par
- The first person on the left is wearing a gray t-shirt with "FITCHBURG STATE" written on it.\par
- The second person is wearing a light blue button-up shirt and khaki pants.\par
- The third person is holding an award and wearing a black dress.\par
- The fourth person is wearing a yellow plaid shirt with gray pants.\par
- The fifth person is wearing a pink jacket over a black blouse and white pants.\par
- The sixth person is wearing a light patterned outfit.\par
\vspace{0.35em}
Step 3: Summing up the total count\par
There are six distinct individuals visible in the image.\par
\vspace{0.35em}
Therefore, the final count is: \textbackslash{}boxed\{6\}\par
\end{tcolorbox}

\clearpage
\subsection{Jigsaw}
\label{app:demo-jigsaw}

\begin{figure}[H]
  \centering
  \begin{tcolorbox}[demoinput={Model Inputs (Tile 1--4)}]
    \centering
    \begin{subfigure}[t]{0.22\linewidth}
      \centering
      \includegraphics[width=\linewidth]{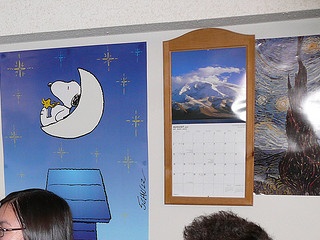}
      \caption*{Tile 1}
    \end{subfigure}\hfill
    \begin{subfigure}[t]{0.22\linewidth}
      \centering
      \includegraphics[width=\linewidth]{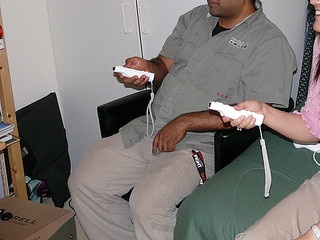}
      \caption*{Tile 2}
    \end{subfigure}\hfill
    \begin{subfigure}[t]{0.22\linewidth}
      \centering
      \includegraphics[width=\linewidth]{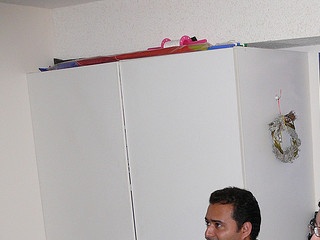}
      \caption*{Tile 3}
    \end{subfigure}\hfill
    \begin{subfigure}[t]{0.22\linewidth}
      \centering
      \includegraphics[width=\linewidth]{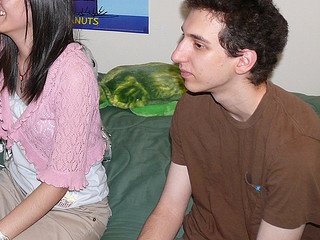}
      \caption*{Tile 4}
    \end{subfigure}
  \end{tcolorbox}

  \vspace{0.45em}
  \begin{tcolorbox}[demoinput={Reconstructed Image (Illustration Only)}]
    \centering
    \includegraphics[width=0.52\linewidth]{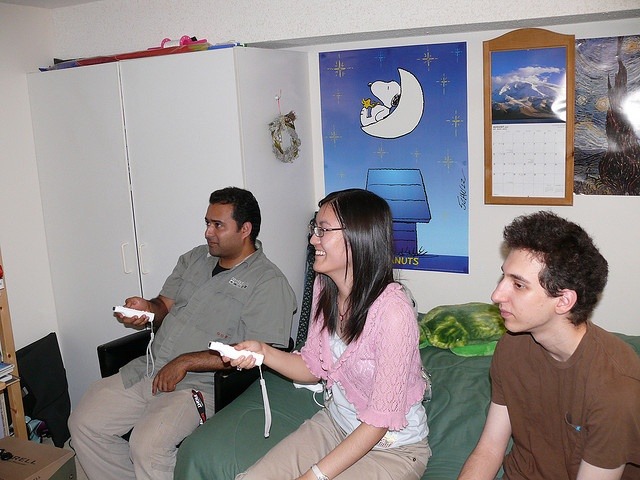}
  \end{tcolorbox}
  \caption{\textbf{A demonstration of Jigsaw inputs and reconstructed image}. Top: the four shuffled tiles fed to the model. Bottom: the original
  photograph reconstructed from the ground-truth ordering.}
  \label{fig:demo-jigsaw}
\end{figure}

Figure~\ref{fig:demo-jigsaw} shows a Jigsaw example.
The four tiles are the model inputs; the reconstructed image is shown only to
aid reading and is not part of the prompt. The unaided rollout proposes the
identity permutation \texttt{[[1, 2], [3, 4]]}. After rewriting the output for revision, the model
recovers the verified ordering \texttt{[[3, 1], [2, 4]]}.

\FloatBarrier

\begin{tcolorbox}[demoexample={Example From Jigsaw}]
\textbf{Problem:}\par\vspace{0.25em}
{\footnotesize\DemoMono
You are given 4 image tiles, labeled Tile 1 to Tile 4 in the order they are presented. These tiles are the four quadrants of one original photograph, cut into a 2x2 grid and then randomly shuffled.\par
\vspace{0.35em}
Your task is to reassemble the original image. Analyze the tiles carefully:\par
- Match colors, textures, lighting and gradients across tile borders.\par
- Trace objects, lines and edges that are cut apart and must continue between adjacent tiles.\par
- Use scene priors: sky / ceiling / distant background usually belongs to the top row, ground / floor / foreground to the bottom row.\par
\vspace{0.35em}
Think step by step, then give the final answer as a 2x2 matrix wrapped in \textbackslash{}boxed\{\}. Row 1 of the matrix is the top row of the original image (left cell first), row 2 is the bottom row. Each cell holds the label number of the tile that belongs at that position.\par
\vspace{0.35em}
For example \textbackslash{}boxed\{[[3, 1], [4, 2]]\} means Tile 3 is the top-left quadrant, Tile 1 the top-right, Tile 4 the bottom-left and Tile 2 the bottom-right.\par
\vspace{0.35em}
Now, write your response as one concise paragraph, no longer than 200 words.\par
}
\vspace{0.6em}
\textbf{Answer:}~\fbox{\texttt{[[3, 1], [2, 4]]}}
\end{tcolorbox}

\begin{tcolorbox}[demowrong]
By analyzing the images, I can see that the top row likely contains the sky/ceiling elements, while the bottom row has ground/floor elements. The first image shows a wall with posters and a calendar, which could be part of the sky/ceiling row. The second image depicts two people holding game controllers, which suggests an indoor setting, possibly the ground/floor. The third image shows a man looking up, and the fourth image shows a person sitting on a bed. The fourth image seems to belong to the ground/floor row due to the bed and the person sitting on it. By matching the colors and textures, the second image also appears to be from the ground/floor row because it features the same type of indoor setting as the fourth image. Thus, we can conclude that the top row consists of the first and second images, and the bottom row consists of the third and fourth images. Therefore, the reassembled image would be \textbackslash{}boxed\{[[1, 2], [3, 4]]\}.\par
\end{tcolorbox}

\begin{tcolorbox}[demoright]
Tile 3 is the top-left because the wardrobe/cabinet background and seated man continue vertically. Tile 1 is the top-right because the cropped heads at the bottom of Tile 1 continue into the people sitting on the bed in Tile 4. Tile 2 is the bottom-left because the man looking up in Tile 2 connects to the seated woman in Tile 4 through the shared edge. Tile 4 is the bottom-right because the person sitting on the bed in Tile 4 is the same as the person in Tile 2. Therefore, the reassembled image would be \textbackslash{}boxed\{[[3, 1], [2, 4]]\}.\par
\end{tcolorbox}

\clearpage
\section{Revision Prompt Templates}
\label{app:demo-templates}

The demonstrations above use a revision stage that edits an incorrect
rollout against an expert reference. Below we list the three task-specific
prompt templates in full.

\subsection{Clock Reading}
\begin{tcolorbox}[demotemplate={Clock Reading Revision Template}]
You are an expert at reading analogue clocks. The dial has three hands: a thin RED second hand, a long BLACK minute hand, and a short thick BLACK hour hand.\par
\vspace{0.35em}
Below are the image, the question, the correct time, a reference answer, and the answer you wrote earlier. Revise your earlier answer so that it is correct, changing as little of it as possible.\par
\vspace{0.35em}
\#\# Rules\par
\vspace{0.35em}
- Make the SMALLEST edit that makes the answer correct. Keep your earlier answer's wording, structure, order of the hands and overall length wherever it is already consistent with the correct time; rewrite only the spans that conflict with it. If a sentence is already right, copy it verbatim.\par
- If your earlier answer is already correct, reproduce it unchanged.\par
- Do not add new sections, headings, summaries, apologies or commentary that your earlier answer did not have, and do not delete parts that were already correct.\par
- Show the reasoning first, hand by hand, then close with the time wrapped in \textbackslash{}boxed\{\}. A reply that jumps straight to the boxed time is rejected.\par
- The time must be exactly \{gt\_time\}, in that digit form (03:45:20 — never 3:45:20, 15:45:20 or 03:45).\par
- Write it as your own reading of the image: never mention the earlier answer, the correct time or the reference answer, and never reuse their wording.\par
\vspace{0.35em}
\#\# Question\par
\vspace{0.35em}
\{question\}\par
\vspace{0.35em}
\#\# Correct time\par
\vspace{0.35em}
\{gt\_time\}\par
\vspace{0.35em}
\#\# Reference answer (style only)\par
\vspace{0.35em}
\{reference\_answer\}\par
\vspace{0.35em}
\#\# Your earlier answer\par
\vspace{0.35em}
\{previous\_output\}\par
\vspace{0.35em}
---\par
\vspace{0.35em}
Output only the revised answer: the complete reading of the dial, keeping as much of your earlier answer as possible, ending with the time in \textbackslash{}boxed\{\}.\par
\end{tcolorbox}

\subsection{Counting}
\begin{tcolorbox}[demotemplate={Counting Revision Template}]
You are an expert visual counting assistant. Your earlier answer reached the wrong total (\{wrong\_count\}). Repair it with the fewest edits that make it genuinely correct.\par
\vspace{0.35em}
You are given the image, the counting question, the verified correct total, a reference answer written by someone else, and your earlier attempt.\par
\vspace{0.35em}
\#\# How to revise\par
\vspace{0.35em}
1. Look at the image again and locate every object the question asks about.\par
2. Compare them against your draft's items: which objects are missing, which are counted twice, which belong to the wrong category, which were merged into a single line.\par
3. Patch exactly those items — add, delete or re-describe them — and renumber if needed.\par
4. Recount your own list. It must account for exactly \{gt\_count\} objects before you finish.\par
\vspace{0.35em}
\#\# Core principle: change as little as possible\par
\vspace{0.35em}
- Keep your draft's opening sentence, item order, numbering and wording wherever they already hold.\par
- Edit only the spans that are wrong; never rewrite, reorder or polish parts that are already fine.\par
- Output the whole draft with the patches applied, not just the lines you changed. If the draft had no enumeration, write one.\par
\vspace{0.35em}
\#\# What "fewest edits" does not mean\par
\vspace{0.35em}
- The enumeration is what went wrong, so the enumeration is what you must repair. Changing only the final number and leaving the list untouched is INVALID.\par
- Your list and your total must agree. If the list accounts for four objects, writing \{gt\_count\} on the last line is a contradiction, not a fix.\par
\vspace{0.35em}
\#\# Do not copy the reference answer\par
\vspace{0.35em}
- The reference answer exists only to show you WHICH objects are in the image. It is a diagnostic aid, never a template.\par
- Never reuse its sentences, phrasing, item order or opening line. Describe each object in your draft's own voice and vocabulary.\par
- Rewriting your answer from the reference wastes the sample: a response that substantially copies it is discarded downstream even when the total is right.\par
\vspace{0.35em}
\#\# Hard requirements\par
\vspace{0.35em}
- Include the full enumeration, one line per object. An answer that is only the final line is rejected.\par
- Every item must describe something actually visible — position, color, appearance or another distinguishing detail. Never invent an object to reach the total.\par
- If objects are too crowded to separate one by one, describe them as a group and state that group's subtotal.\par
- Write as if you had read the image yourself: never mention the reference answer, the verified total, your earlier attempt, or this instruction block.\par
- Close with the total on its own last line, in exactly this form: Final count: \textbackslash{}(\textbackslash{}boxed\{N\}\textbackslash{})\par
\vspace{0.35em}
\#\# Counting question\par
\vspace{0.35em}
\{question\}\par
\vspace{0.35em}
\#\# Verified correct total\par
\vspace{0.35em}
\{gt\_count\}\par
\vspace{0.35em}
\#\# Reference answer (to identify the objects only; never quote it)\par
\vspace{0.35em}
\{reference\_answer\}\par
\vspace{0.35em}
\#\# Your earlier attempt (the draft to edit; it answered \{wrong\_count\})\par
\vspace{0.35em}
\{wrong\_output\}\par
\vspace{0.35em}
---\par
\vspace{0.35em}
Output only the repaired answer: your edited draft, with the enumeration lines followed by the final count line. No preamble, no notes on what you changed.\par
\end{tcolorbox}

\subsection{Jigsaw}
\begin{tcolorbox}[demotemplate={Jigsaw Revision Template}]
You are an expert at reassembling shuffled image tiles. You are given the four tiles of one photograph, the question that was asked about them, the verified correct arrangement, a reference answer, and a previous INCORRECT attempt made by a model.\par
\vspace{0.35em}
Your task is to REPAIR that attempt: fix what is wrong in its reasoning and in its final answer, while keeping whatever it already got right. Whatever the attempt looks like, your own answer must always be a complete step-by-step derivation — never a bare matrix.\par
\vspace{0.35em}
\#\# Required output shape\par
\vspace{0.35em}
Output a full, self-contained solution that reasons step by step and ends with the final matrix:\par
\vspace{0.35em}
1. One step per position: say which tile goes in the top-left, the top-right, the bottom-left and the bottom-right, and for each one give the specific visual evidence that puts it there (which object, edge, line, horizon or texture is cut apart and continues into which neighbouring tile, or which scene cue such as sky/distant background vs ground/foreground fixes its row).\par
2. One step that confirms the left/right order inside each row, since a row can be right while its two tiles are swapped.\par
3. The final line, in exactly this form:\par
\vspace{0.35em}
Therefore the correct reconstruction is \textbackslash{}(\textbackslash{}boxed\{[[a, b], [c, d]]\}\textbackslash{})\par
\vspace{0.35em}
\#\# Repair the reasoning — never append to it\par
\vspace{0.35em}
The previous attempt reached the wrong arrangement because parts of its reasoning are wrong. Fix that reasoning; do not work around it.\par
\vspace{0.35em}
- Correct every claim that led to the wrong answer. A sentence that puts a tile in the wrong quadrant, that says a seam, object or texture continues where it does not, or that concludes a row/column order disagreeing with the final matrix, must be rewritten — not kept.\par
- NEVER keep the attempt's wrong conclusion and bolt the right answer onto the end. Appending a line like "Therefore the correct reconstruction is ..." after the attempt's original wrong conclusion is the worst possible outcome: it leaves the answer arguing for one arrangement while stating another.\par
- Your output must contain EXACTLY ONE \textbackslash{}boxed\{\}, the final one. Remove the attempt's old matrix; never leave two matrices in the answer.\par
- Do not just copy the attempt's text and change the digits in the last matrix. If the reasoning above the matrix still supports the old arrangement, the answer is wrong even though the matrix is right, and it will be rejected.\par
- Before you finish, read your answer back once and check it against \{gt\_readable\}: every quadrant you name, every piece of evidence and the final matrix must agree with each other and with that arrangement. Fix anything that does not.\par
\vspace{0.35em}
\#\# Hard requirements\par
\vspace{0.35em}
- THE STEP-BY-STEP REASONING IS MANDATORY, no matter what the previous attempt looks like. An answer that is only the final matrix, or that has a couple of sentences with no visual evidence, is INVALID and will be rejected. Always write all four position steps plus the left/right check.\par
- The final matrix must be exactly \{gt\_matrix\}, i.e. \{gt\_readable\}.\par
- When the previous attempt has no reasoning to preserve — for instance it is a bare matrix — there is simply nothing to keep: derive the whole answer yourself from the tiles. Answering with another bare matrix is the worst possible output. When it does have reasoning, your answer must be at least as detailed as it is.\par
- Keep what already works: reproduce word for word those sentences of the previous attempt that are consistent with the correct arrangement, and preserve its wording, order and style. Change what is wrong, and nothing else.\par
- Name the tiles explicitly as "Tile 1" ... "Tile 4" and cite only evidence actually visible in the tiles — never invent an object, an edge or a marking, and never pad the text to make it longer.\par
- State each tile's position the way the final matrix does. Every "Tile N is the top-left / top-right / bottom-left / bottom-right" claim in your text must match \{gt\_readable\} exactly; a single mismatched quadrant makes the whole answer invalid.\par
- DO NOT COPY THE REFERENCE ANSWER, and do not reuse any wording from these instructions. The reference answer is shown only so you can check which visual evidence is true; where it and the previous attempt say the same thing differently, keep the previous attempt's wording.\par
- Write as if the attempt had been right the first time. NEVER mention the reference answer, the verified arrangement, the previous attempt or the fact that anything was corrected; avoid words such as "reference", "ground truth", "hint", "the given arrangement", "previously" and "instead of".\par
\vspace{0.35em}
\#\# Jigsaw Question\par
\vspace{0.35em}
\{question\}\par
\vspace{0.35em}
\#\# Verified Correct Arrangement\par
\vspace{0.35em}
\{gt\_matrix\}  (\{gt\_readable\})\par
\vspace{0.35em}
\#\# Reference Answer (known correct; consult it ONLY to check which evidence is true — never copy from it)\par
\vspace{0.35em}
\{reference\_answer\}\par
\vspace{0.35em}
\#\# Model's Incorrect Attempt (it answered \{wrong\_matrix\}) — repair this text: correct its faulty reasoning, keep its correct parts, and do not merely append an answer to it\par
\vspace{0.35em}
\{wrong\_output\}\par
\vspace{0.35em}
---\par
\vspace{0.35em}
Output ONLY the repaired answer itself: the step-by-step derivation covering all four positions and the left/right check, ending with a single `\textbackslash{}boxed\{\}` line whose reasoning fully supports it. No preamble, no diff, no list of what you changed.\par
\end{tcolorbox}

\clearpage
\section{Concise Revision Prompt Templates}
\label{app:revision-prompt}

Below we present more concise and abstract revision templates, without the fine-grained
rule constraints of Appendix~\ref{app:demo-templates}. Empirically, this style of prompt
underperforms on smaller models (3B and 7B) but works well on larger ones (32B and 72B),
indicating that revision competence improves with model scale.

\subsection{Clock Reading}
\begin{tcolorbox}[demotemplate={Clock Reading revision template (concise)}]
You are an expert at reading analogue clocks. The dial has three hands: a long, thin RED second hand, a long BLACK minute hand, and a short, thick BLACK hour hand.\par
\vspace{0.35em}
Below are the image, the question, a reference reasoning trajectory, and a model's WRONG attempt. Compare the wrong attempt with the correct reference trajectory, identify the specific mistakes in the wrong attempt, and revise it to fix those mistakes. Your task is only to repair the wrong attempt. Use the reference trajectory solely as a source of information for correcting errors; do not copy it as a replacement for the wrong attempt.\par
\vspace{0.35em}
\#\# Rules\par
\vspace{0.35em}
- The time must be written in the zero-padded 12-hour HH:MM:SS format (for example, 03:45:20 — never 3:45:20, 15:45:20, or 03:45).\par
- Correct the specific mistakes made in the wrong attempt.\par
\vspace{0.35em}
\#\# Question\par
\vspace{0.35em}
\{question\}\par
\vspace{0.35em}
\#\# Reference trajectory\par
\vspace{0.35em}
\{reference\_answer\}\par
\vspace{0.35em}
\#\# Wrong attempt\par
\vspace{0.35em}
\{wrong\_output\}\par
\vspace{0.35em}
---\par
\vspace{0.35em}
Now write the repaired version of the wrong attempt. Minimize the edit distance from the original attempt: preserve its wording and structure as much as possible, changing only what is necessary to correct the errors.\par
\end{tcolorbox}

\subsection{Jigsaw}
\begin{tcolorbox}[demotemplate={Jigsaw revision template (concise)}]
You are an expert at reassembling shuffled image tiles. The four tiles above are the quadrants of one photograph, cut into a 2x2 grid and then shuffled.\par
\vspace{0.35em}
Below are the tiles, the question, a reference answer, and a model's WRONG attempt. Compare the wrong attempt with the correct reference answer, identify the specific mistakes in the wrong attempt, and revise it to fix those mistakes. Your task is only to repair the wrong attempt. Use the reference answer solely as a source of information for correcting errors; do not copy it as a replacement for the wrong attempt.\par
\vspace{0.35em}
\#\# Rules\par
\vspace{0.35em}
- The final answer must be a 2x2 matrix wrapped in \textbackslash{}boxed\{\}, written as \textbackslash{}(\textbackslash{}boxed\{[[a, b], [c, d]]\}\textbackslash{}), and your answer must contain no other matrix.\par
- The wrong attempt put the tiles in the wrong order, so the sentences that placed them must change too. Rewrite every sentence that puts a tile in a quadrant it does not occupy, or that claims something continues between tiles that are not neighbours. Never leave a sentence arguing for one arrangement while your matrix states another.\par
- Apart from those sentences, change as little as possible: keep the attempt's own wording, sentence order, level of detail and overall length.\par
- Present it as your own reading of the tiles: never mention the attempt or the reference answer, and do not say that anything was corrected.\par
\vspace{0.35em}
\#\# Question\par
\vspace{0.35em}
\{question\}\par
\vspace{0.35em}
\#\# Reference answer\par
\vspace{0.35em}
\{reference\_answer\}\par
\vspace{0.35em}
\#\# Wrong attempt\par
\vspace{0.35em}
\{wrong\_output\}\par
\vspace{0.35em}
---\par
\vspace{0.35em}
Now write the repaired version of the wrong attempt. Minimize the edit distance from the original attempt: preserve its wording and structure as much as possible, changing only what is necessary to correct the errors.\par
\end{tcolorbox}

\end{document}